\documentclass[preprint,12pt]{elsarticle}

\usepackage{multirow}
\usepackage{amssymb}
\usepackage{amsmath}
\usepackage{array}

\journal{Nuclear Physics B}

\begin{document}

\begin{frontmatter}



\title{Clinically Interpretable Sepsis Early Warning via LLM-Guided Simulation of Temporal Physiological Dynamics}

\cortext[mycor]{Corresponding authors: Chunpei Li (licp@gxnu.edu.cn), Ke Lu (sgu8434@sina.com) and Hongzhi Yu}

\affiliation[inst1]{organization={School of Electrical and Information Engineering, Tianjin University},%
            city={Tianjin},
            postcode={300072}, 
            state={Tianjin},
            country={China}}
\affiliation[inst2]{organization={Department of Information Engineering and Computer Science, University of Trento},%
            city={Trento},
            state={Trentino},
            country={Italy}}
\affiliation[inst3]{organization={Department of Orthopedics, Affiliated Kunshan Hospital of Jiangsu University},%
            city={Kunshan},
            state={Jiangsu},
            country={China}}
\affiliation[inst4]{organization={Tianjin University Chest Hospital},%
            city={Tianjin},
            postcode={300072}, 
            state={Tianjin},
            country={China}}
\affiliation[inst5]{organization={Haihe Hospital, Tianjin University},%
            city={Tianjin},
            postcode={300072}, 
            state={Tianjin},
            country={China}}
\affiliation[inst6]{organization={Key Laboratory of Education Blockchain and Intelligent Technology, Ministry of Education, Guangxi Normal University},%
city={Guilin},
postcode={541004},
state={Guangxi},
country={China}}

\author[inst1]{Weizhi Nie}

\author[inst1]{Zhen Qu}

\author[inst2]{Weijie Wang}

\author[inst6]{Chunpei 
Li\corref{mycor}}

\author[inst3]{Ke Lu\corref{mycor}}

\author[inst4]{Bingyang Zhou}

\author[inst5]{Hongzhi Yu\corref{mycor}}

\begin{abstract}
Timely and interpretable early warning of sepsis remains a major clinical challenge due to the complex temporal dynamics of physiological deterioration. Traditional data-driven models often provide accurate yet opaque predictions, limiting physicians’ confidence and clinical applicability. To address this limitation, we propose a Large Language Model (LLM)-guided temporal simulation framework that explicitly models physiological trajectories prior to disease onset for clinically interpretable prediction. The framework consists of a spatiotemporal feature extraction module that captures dynamic dependencies among multivariate vital signs, a Medical Prompt-as-Prefix module that embeds clinical reasoning cues into LLMs, and an agent-based post-processing component that constrains predictions within physiologically plausible ranges. By first simulating the evolution of key physiological indicators and then classifying sepsis onset, our model offers transparent prediction mechanisms that align with clinical judgment. Evaluated on the MIMIC-IV and eICU databases, the proposed method achieves superior AUC scores (0.861–0.903) across 24–4-hour pre-onset prediction tasks, outperforming conventional deep learning and rule-based approaches. More importantly, it provides interpretable trajectories and risk trends that can assist clinicians in early intervention and personalized decision-making in intensive care environments.
\end{abstract}

\begin{keyword}
Sepsis, Early warning system, Clinical interpretability, LLM-based modeling, Temporal simulation, MIMIC-IV.
\end{keyword}

\end{frontmatter}



\section{Introduction}
\label{sec:introduction}

\begin{figure}[htp]
    \centering
    \includegraphics[width=7cm]{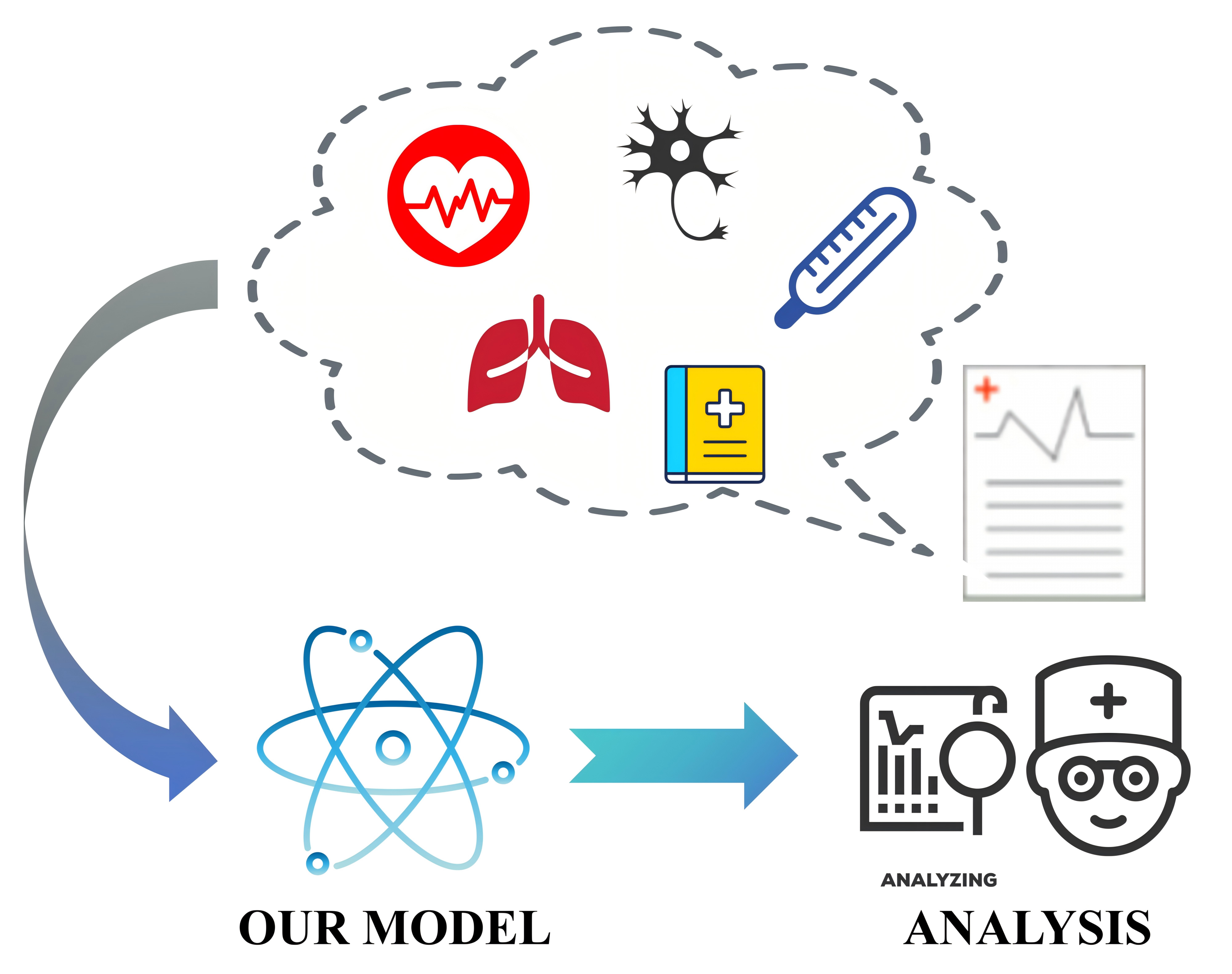}
    \caption{Illustration of the LLM-guided “predict-then-classify” mechanism for sepsis early warning.
The model first simulates key physiological indicators through temporal reasoning, then classifies sepsis onset based on the simulated trajectories.
This design provides clinically interpretable predictions that reveal how and why physiological deterioration occurs before diagnosis.}
    \label{fig:result}
\end{figure}
As a life-threatening systemic inflammatory syndrome, sepsis is characterized by organ dysfunction resulting from a dysregulated host response to infection. It has worldwide become the leading cause of mortality in intensive care units (ICUs) \cite{bib1}. Moreover, it has been recognized as a global health priority because of its significant clinical complexity and heterogeneity \cite{bib2}. According to World Health Organization statistics, sepsis causes millions of deaths annually. Its incidence shows a persistent upward trajectory. Some studies have demonstrated that each hour of delay in antibiotic administration corresponds to a measurable increase in mortality risk \cite{bib3}. All of these can demonstrate the critical importance of timely intervention. Consequently, an effective early warning system can secure crucial time for clinical decision-making, proving to be decisive for patient outcomes.

Currently, the Sequential Organ Failure Assessment (SOFA) score \cite{bib4} has been widely used in clinical practice. For each physiological parameter, it is based on the worst evaluation values within 24 hours. While this scoring system can effectively capture disease progression, it still relies on the extreme values of a single time point. Besides, its capacity to utilize dynamic information embedded in time-series data is limited by the lack of historical temporal patterns. Therefore, the SOFA score has substantial constraints for early warning purposes.

People attempt to develop an early warning system for sepsis prediction. Most sepsis prediction models predominantly use multivariate time-series data \cite{bib8} to predict whether a patient will develop sepsis. The result is often a binary label, lacking certain interpretability. Therefore, our goal is to first predict the numerical values of physiological indicators and then classify. Additionally, existing models typically incorporate patient demographics, laboratory results, and physiological indicators\cite{bib62}. However, sepsis has complex pathophysiology, involving intricate interactions across multiple physiological systems and signaling pathways \cite{bib36}. Therefore, we need to focus on not only analyzing the spatiotemporal features of structured data, but also considering the patients’ situations before onset and medication histories. The emergence of the large language models(LLMs) helps solve this problem. Some researchers have begun exploring LLMs as auxiliary tools for sepsis prediction. Because of their excellent text recognition capability, LLMs show great potential in sepsis prediction.

\subsection{Motivation}
Researchers have comprehensively analyzed clinical text data, physiological parameters, and biosignals. Based on these analyses, we have developed precise early prediction models for sepsis. Generally, patients’ physiological indicators are captured as time-series data. By applying fixed-length observation windows, we can extract time-aligned data segments. Additionally, patients’ individualized textual descriptions also provide valuable contextual information.

Our research aims to enhance the accuracy of early sepsis prediction models. The key challenge is how to effectively utilize time-series data and textual information to improve prediction \cite{bib37}. LLMs have demonstrated their significant advantages in processing unstructured textual data. Because of their inherent Transformer architecture, they also show potential for analyzing temporal sequences. Therefore, we must consider the following three critical issues when using LLMs to address sepsis prediction:

\begin{itemize}
  \item How to extract effective and credible feature vectors from sparse patient data for prediction? This requires the module to comprehensively mine structured patient data. Moreover, it is also crucial to enable LLMs to properly interpret the clinical significance.
  \item How to improve LLMs' early-warning capability for accuracy prediction? We need to develop more effective prompt strategies to improve LLMs' learning performance.
  \item How to improve the interpretability of final predictions? The predicted time-series data can not only provide strong support in clinical diagnosis but also improve the accuracy of the subsequent classification process.
\end{itemize}

In order to address these challenges, we designed a novel LLM-based spatiotemporal feature extraction module. Our model effectively integrates textual data and time-series data. Furthermore, our model outputs the predicted value of the onset to enhance the interpretability of the results. Through experiments, our module can not only uncover latent relationships in the data but also transform them into interpretable features for LLMs. Additionally, we employ medical prompt engineering to enhance model comprehension ability. To further improve the accuracy of prediction results, we designed an agent-based post-processing module to constrain the results. This module performs prior predictions before integrating the optimized outputs with the raw data for final classification.

\subsection {Contribution}

Our framework holistically integrates diverse patient data, including clinical notes and time-series data. By using the prior prediction module, our model not only improves prediction accuracy but also delivers more reliable and interpretable predicted data. It can provide stronger support for clinical insights. Contributions are followed as:
\begin{itemize}
  \item  We used an LLM-based Series Data Prediction module for early sepsis prediction. It predicts patients’ time-series data and uses simulated results to enhance classification accuracy. In our experiments, the module demonstrated superior predictive performance.
  \item We used LLMs to guide and constrain prediction results. Through an agent-based post-processing module, it can prevent predictions from exceeding normal ranges, improving prior prediction precision, reliability, and explaining final outcomes.
  \item We used two different datasets and defined sepsis onset as the time point when a patient's SOFA score reaches or exceeds 2 in data processing. After filtering, we used fixed-duration patient data and tested with another dataset to verify the model. Data processing details are shown in the next section.
\end{itemize}

\section{DATA PROCESSING}

The database used in our research is the Medical Information Mart for Intensive Care-IV (MIMIC-IV v2.2) \cite{bib12}. This publicly available medical database is derived from the electronic health record system of Beth Israel Deaconess Medical Center, containing the inpatient information from 2008 to 2019. Compared with version 2.0, MIMIC-IV v2.2 provides access to discharge summaries and radiology reports, while retaining the original structured data. It adopts a modular design, including hosp, icu, and note modules. Moreover, the database has performed privacy protection processing. Each patient has a unique identifier as "subject\_id" and each hospitalization record also has a sole "hadm\_id". By utilizing these identifiers, researchers can build cross-table data associations, and accurately extract target data.

\subsection {Extracting Initial Samples}
At the beginning of the research, we screened potentially relevant data from the database. The total processing flow is shown in Figure. 2. The only thing we need to focus on is extracting ICU patients’ data from the MIMIC-IV database for early sepsis prediction. We only utilized the patient data collected after their admission to the ICU. Based on clinical criteria, we selected to define sepsis onset as the time point when their SOFA score reached $\geq$2. Furthermore, we initially extracted all available ICU admission records. To better screen the patient dataset, we utilized the following criteria: (1) patients aged $\geq$18 years during their first hospitalization. (2) each hospitalization had only a single ICU transfer.
After the initial screening process, the final dataset included 42372 patients and 55452 ICU admissions. In subsequent analyses, each ICU admission record was used as a basic analysis unit, rather than treating individual patients as independent samples.

\begin{figure}[htp]
    \centering
    \includegraphics[width=8cm]{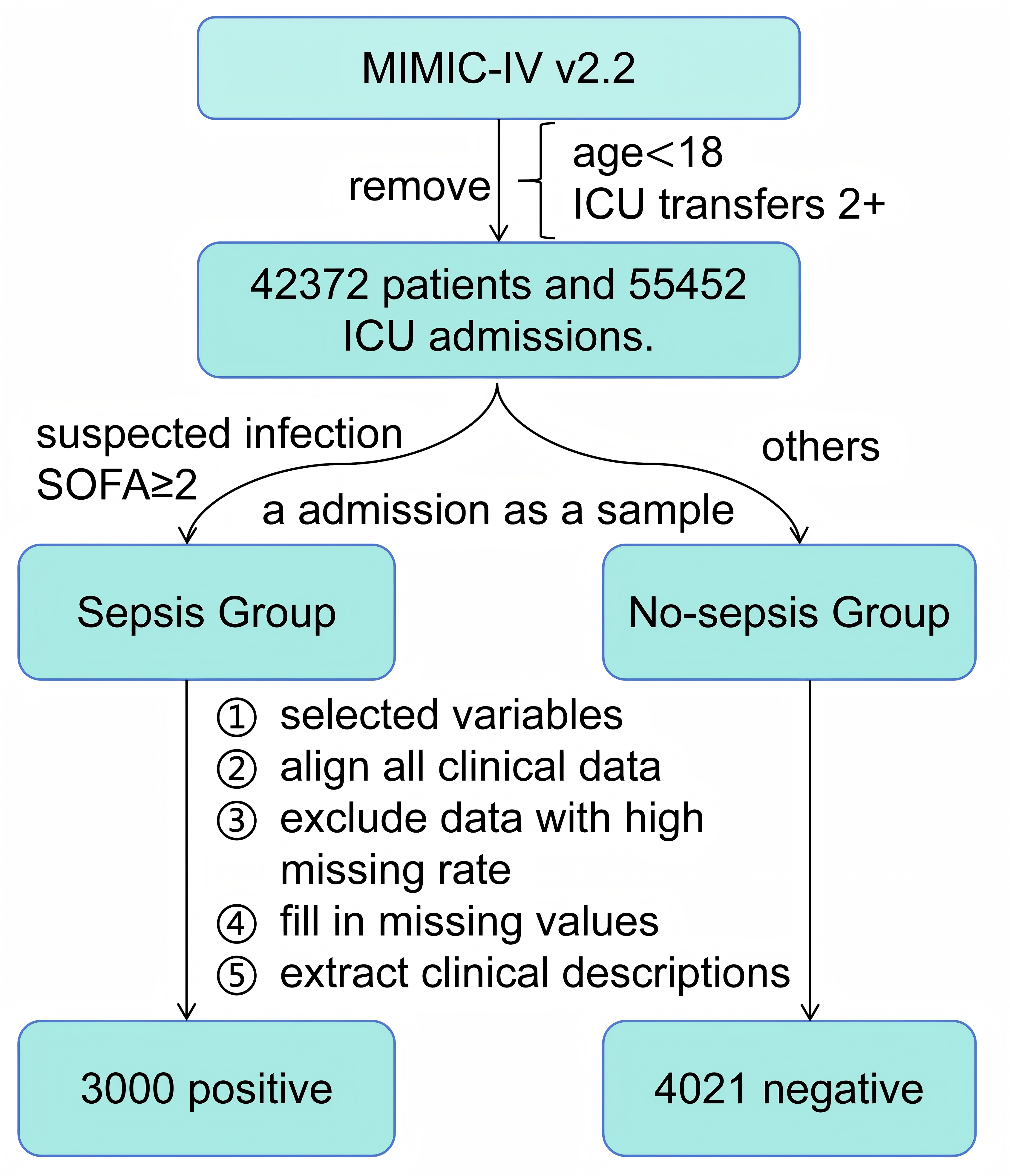}
    \caption{Flowchart of data processing. The above figure illustrates the screening criteria and operational procedures in the patient data processing. We finally got 3000 positive samples and 4021 negative samples.}
    \label{fig:result}
\end{figure}

\subsection {Positive and Negative Samples}
In this research, we utilized three sepsis diagnostic categories defined in the International Classification of Diseases (ICD) coding system, including sepsis, severe sepsis, and septic shock \cite{bib13}. Based on the clinical assessment indicators, we labeled cases with SOFA scores greater than 2 as positive samples, while the others were negative samples. The final dataset contained 24,100 positive samples and 31,352 negative samples. As for the method of data processing, we compared two temporal alignment approaches:

\begin{itemize}
  \item Left-alignment using ICU admission time. Based on the time of ICU admission, clinical features were extracted at predefined fixed intervals, such as 12, 18, and 24 hours. While this method ensures comparison of these samples with the same hospitalization duration, the real onset time point of sepsis varies among individuals. A representative application is the dynamic monitoring approach utilized by Goh et al. \cite{bib14}. This method employed multiple time-point data after ICU admission.
  \item Right-alignment based on disease progression. Based on the time of diagnosis for sepsis, clinical data were uniformly extracted from the fixed window before the onset \cite{bib15}. This method can ensure all of the features extracted from positive samples are consistent with the onset time of the disease. While negative samples lack clear event markers, those samples would be aligned using relaxed criteria.
\end{itemize}

We fully compared the two methods. After evaluating our model’s objectives and performance requirements, we selected to adopt the right-alignment strategy. We designated the first occurrence of SOFA scores $\geq$ 2 as the sepsis onset marker. By this method, our dataset can better reflect the dynamic pathological progression of sepsis.

\subsection {Data Cleaning}
To address data quality issues in the raw dataset, we systematically implemented the preprocessing pipeline like Figure. 2. With the guidance of local hospital medical experts, we first selected 47 physiological variables, including both static patient information and dynamic measurements, and summarized them into a structured table. After that, due to the complexity of the data time points, we selected to align all clinical data at hourly intervals. As for those variables with multiple measurements within the same hour, we only used the latest value. Sometimes clinicians do not make records about those parameters within normal ranges. To ensure prediction reliability, we consulted medical experts and excluded patient records with a higher rate of absence and outliers \cite{bib53}.

For missing values, we selected to utilize the method of searching forward for the nearest available observation, then backward if necessary \cite{bib18}. Moreover, if some variables are completely missing, we fill in these values by using the median value within the clinically normal range. Meanwhile, to leverage the text analysis capabilities of LLMs, we needed to extract personalized clinical descriptions from each patient's electronic health records. We utilized the Deepseek model to summarize diagnostic texts from hospitalization records, including admission assessments, surgical reports, and medication regimens. The final organized dataset contained 3,000 positive samples and 4,021 negative, with a time span of 30 hours prior to the onset of disease. Training:testing: validation sets were allocated in a 7:2:1 ratio.

\section{METHOD}

\subsection {Problem Definition}

\begin{figure*}[htp]
    \centering
    \includegraphics[width=13cm]{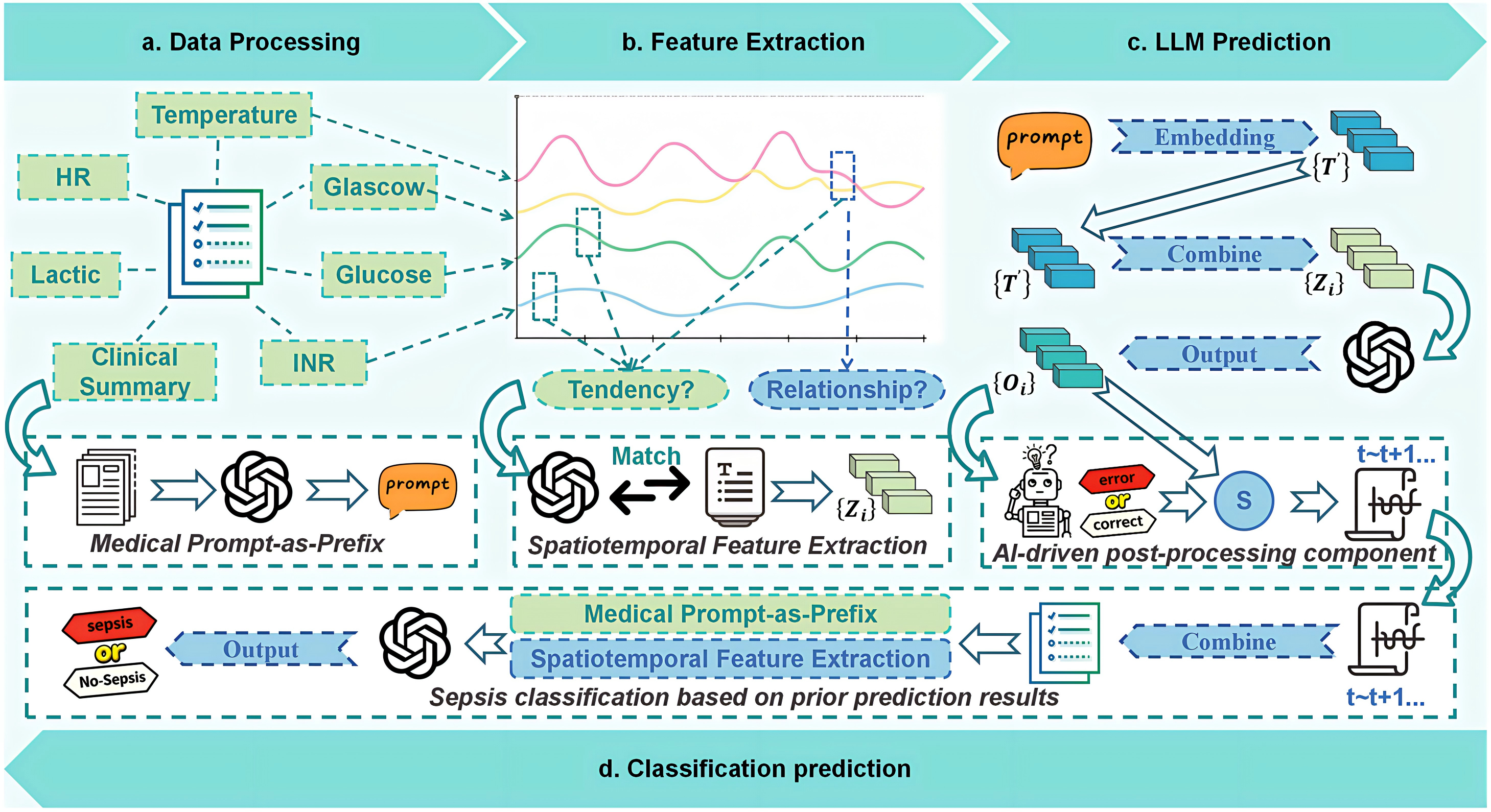}
    \caption{The operational mechanism of our proposed LLM-Based Spatiotemporal Feature Extraction Model is as follows: First, the patient data is processed using an LLM to extract and organize information. The patients’ clinical summaries are then utilized through the Medical Prompt-as-Prefix module. Subsequently, spatiotemporal feature extraction is performed on the time-series data. Specifically, the LLM's vocabulary and the time-series data are used to conduct two rounds of correlation feature extraction along the temporal and spatial dimensions, respectively. This step aims to derive spatiotemporal features that the LLM can comprehend, which are then aligned and fused with the vectorized representations. The combined features are fed into a partially frozen LLM for prediction. To enhance the model's predictive performance and interpretability, an AI-driven post-processing component is employed to output physiological indicators at the onset time $t$. These indicators are integrated with the original data and re-input into the model for classification prediction. Finally, a classifier is used to predict the occurrence of sepsis.}
    \label{fig:model}
\end{figure*}

In this section, we show how to construct a spatiotemporal feature extraction model that enables LLMs to comprehend and perform predictions. 
In Figure. 3, we first define the processed patient data as $\mathbf{X} \in \mathbb{R}^{N \times T}$. This variable means that $N$ distinct dynamic physiological indicators span $T$ time steps. We utilize all patient data prior to time $t$, which is the time point we need to predict. Our aim is to transform this data through a function $f(\cdot)$ into time-series features $\mathbf{Z}$ that are interpretable by the LLMs.
Next, we need to guide LLMs to understand time-series data and the prediction task \cite{bib39}. Therefore, we design a Medical Prompt-as-Prefix module. By utilizing LLMs to generate textual descriptions $\mathbf{T}$, we can obtain the patient summary, containing trend patterns and early warning signals from the patient data. Adding a task statement, $\mathbf{T}$ is subsequently vectorized into $\mathbf{\hat{T}}$ through prompt engineering.
The model will integrate the prompt embeddings $\mathbf{\hat{T}}$ with feature vectors $\mathbf{Z}$ to input into a partly parameter-frozen LLM. By comprehensive analysis, LLM will predict future physiological indicators $\mathbf{X}_t$ at the onset time.
Finally, these predicted values $\mathbf{X}_t$ are then combined with historical indicators prior to time $t$ to collectively predict the occurrence status at time $t$.

\subsection {Spatiotemporal Feature Extraction}

To enable the LLMs to better comprehend the structured patient data, we designed a spatiotemporal feature extraction module. The aim of this module is to segment the structured patient data and align it with the text prototypes used by LLMs. Furthermore, this module can connect time-series data with natural language data across modalities, activating the backbone model’s capability in understanding and reasoning about temporal patterns.

Concretely, we first employ a sliding window approach to partition the data $\mathbf{X}^{(i)}$ into multiple consecutive segments, where each segment has a length of $L_p$. The total number of input segments $P$ can be expressed as:

\begin{equation}
P = \left\lfloor\frac{T - L_p}{S}\right\rfloor + 2,
\end{equation} 
where $S$ denotes the horizontal sliding step size. This design is primarily motivated by two considerations: (1) better preservation of local semantic information through aggregation within each segment. (2) efficient processing of the LLM's vocabulary $E$. 

Direct utilization of $E$ would lead to an excessively large and potentially dense reprogramming space. As a solution, we maintain a small set of text prototypes through linear probing of $E$. Specifically, we employ a compact linear layer to transform the vocabulary $E$ into text prototypes $\mathbf{E}' \in \mathbf{R}^{V' \times D}$, where $V' \ll V$.

After processing both the data and text prototypes, we employ multi-head attention mechanisms to establish correspondence between them. For each attention head $k = \{1, \ldots, K\}$, we define the respective query matrix $\mathbf{Q}^{(i)} = \hat{\mathbf{X}}_P^{(i)} \mathbf{W}_k^Q$, key matrix $\mathbf{K}_k^{(i)} = \mathbf{E}' \mathbf{W}_k^K$, and value matrix $\mathbf{V}_k^{(i)} = \mathbf{E}' \mathbf{W}_k^V$. The reprogramming operation for each attention head is then defined as:

\begin{equation}
\mathbf{Y}_k^{(i)} = \text{ATTENTION}\left(\mathbf{Q}_k^{(i)}, \mathbf{K}_k^{(i)}, \mathbf{V}_k^{(i)}\right).
\end{equation}

By aggregating the outputs $\mathbf{Y}_k^{(i)}$ from all attention heads, the model gains preliminary recognition of basic data patterns such as "lactate level increase" or "blood pressure decrease" and obtains a hidden state $\mathbf{H}$. However, for sepsis prediction, inter-variable correlations cannot be overlooked\cite{bib50}. Therefore, after obtaining $\mathbf{Y}^{(i)}$, we employ a Transformer architecture to query correlations among variables:

\begin{equation}
\mathbf{Z}_k^{(i)} = \text{ATTENTION}\left(\mathbf{H} \mathbf{W}_k^Q, \mathbf{H} \mathbf{W}_k^K, \mathbf{H} \mathbf{W}_k^V\right).
\end{equation}

Following aggregation and linear projection to align with the backbone model's hidden dimensions, we ultimately obtain $\mathbf{Z}^i \in \mathbf{R}^{P \times D}$. In other words, this step enables the LLM to recognize relationships such as "the coordinated increase of respiratory rate and heart rate with clinical outcomes."

\subsection {Medical Prompt-as-Prefix}
Prompt engineering serves as a direct and effective method for task-specific activation of LLMs \cite{bib34}. However, there are two significant challenges that emerge in this task: First, it has difficulties in directly translating time-series data into natural language representations, which hinders the construction of instructions and constrains the effectiveness of real-time prompting. Second, to integrate medical domain knowledge into LLMs, we require careful consideration for specific tasks. We draw on previous research to address these problems. Finally, we design a Medical Prompt-as-Prefix module to enrich the input context and enhance the capability of LLMs in recognizing medical time-series patterns.

We identify three critical components for constructing effective prompts: 

\begin{itemize}
  \item [(1)]
\textit{Dataset context}: Providing background information about the input time-series data and guiding the LLM to perform the designated task (classification or regression). 
  \item [(2)]
\textit{Key variable alerts}: Based on the criteria of sepsis evaluation, we select each patient’s physiological variables and define their risk thresholds. When these measurements exceed the thresholds, the system generates warning signals. 
  \item [(3)]
\textit{Input statistics}: By analyzing the trend patterns and differential characteristics of physiological parameters\cite{bib33}, LLMs can generate a statistical description of these time-series data. This method is helpful in pattern recognition and clinical reasoning. A representative prompt example is illustrated in Figure. 4.
\end{itemize}

\begin{figure}[htp]
    \centering
    \includegraphics[width=12cm]{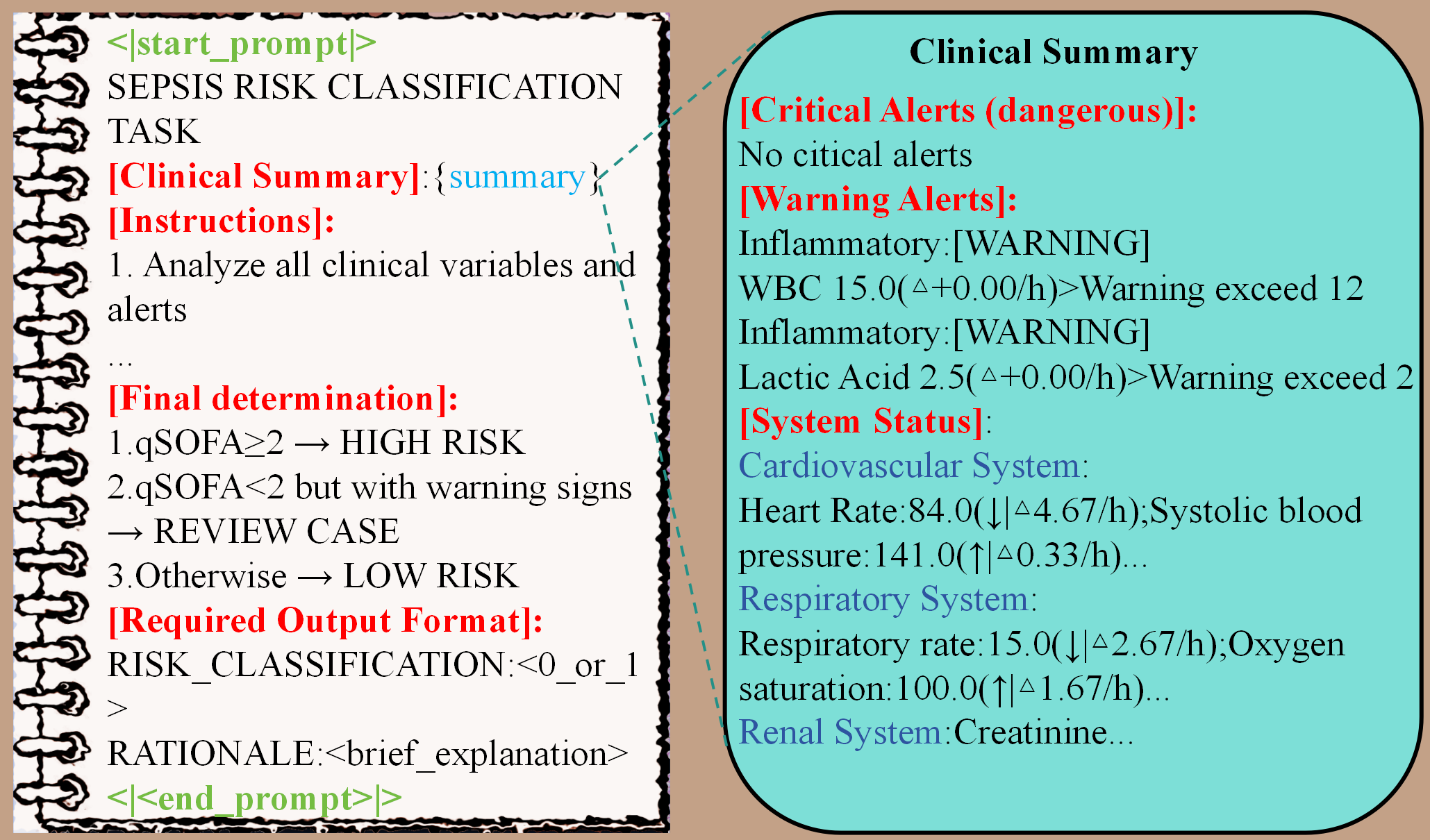}
    \caption{Demonstration of Patient-Level Prompts. The LLM initially performs a comprehensive summarization and refinement of the patient's time-series data, generating an overview of temporal trends and risk assessments while issuing Warning alerts for abnormal data points to produce a final textual patient summary. Subsequent prompt engineering incorporates this patient summary to provide sepsis diagnostic criteria and explicitly specifies the task type, thereby enhancing the LLM's predictive output accuracy.}
    \label{fig:prompt}
\end{figure}

\subsection {AI-driven post-processing component}

There are two stages in the overall model process. In the first stage, the model predicts physiological indicators at time $t$. Specifically, following spatiotemporal feature extraction, $\mathbf{Z}^i$ undergoes a temporal global average pooling, and produces an output vector $\mathbf{Z} \in \mathbf{R}^{N \times D}$. This vector is concatenated with the prompt embeddings and fed into the partially frozen LLM. Furthermore, we get the output representations after removing the prefix tokens from the LLM’s output. These representations are then flattened and projected through a linear layer. After that, we obtain the preliminary prediction results $\mathbf{O}_t$.

To enhance prediction accuracy, we design an agent-based post-processing module. Specifically, we first provide the LLM with validated normal ranges and variation amplitudes for all target variables. The aim is to enable the LLM to detect predicted outliers. Upon receiving the initial predictions $\mathbf{O}_t$, the LLM first evaluates whether they exceed the range we provide. If the prediction is normal, the LLM does not adjust. For out-of-range predictions, the LLM generates correction methods based on the degree of deviation. The model then utilizes these corrections through a lightweight linear layer. Finally, this layer will learn scaling parameters $\mathbf{s} \in \mathbf{R}^{N}$, which are used to constrain outliers and produce the final adjusted outputs $\mathbf{X}_t$.

\subsection {Classification output}

To ensure optimal model performance, we perform classification and regression tasks separately. We utilize the outcomes of regression to enhance the predicted performance of classification \cite{bib54}. In the second stage, the model first concatenates the predicted physiological indicators with historical data up to time $t-1$. After that, this combined dataset is then reprocessed through the model for classification training. At this stage, the model uses all available data up to and including time $t$ for prediction. Through a two-layer multilayer perceptron (MLP) classification head, we obtain the final outcomes.

\begin{equation}
\hat{y} = \text{MLP}_2(\sigma(\text{MLP}_1(\mathbf{h}_t))),
\end{equation}
where $\mathbf{h}_t$ represents the LLM's hidden states at time $t$, $\sigma$ denotes the activation function, and $\text{MLP}_1$, $\text{MLP}_2$ constitute the two fully-connected layers of the classification head. This dual-stage architecture enables more effective optimization of the classified output data, while maintaining the performance requirements.

\section{EXPERIMENT}
\subsection {Evaluation}

In this section, we use various evaluation metrics, depending on the task outcome. For example, early sepsis prediction is inherently a binary classification task. Therefore, the primary evaluation metric we use is the AUC-ROC (Area Under the Receiver Operating Characteristic Curve). For this metric, a higher AUC value indicates better predictive performance of the model. To further validate the results predicted by our model, we use the F1-score, true positive rate (TPR), and false positive rate (FPR) as supplementary evaluation metrics. For the prior prediction module’s results, we adopt the mean squared error (MSE) as the metric to assess prediction accuracy.

\subsection {Results}

\begin{figure}
    \centering
    \includegraphics[width=8cm]{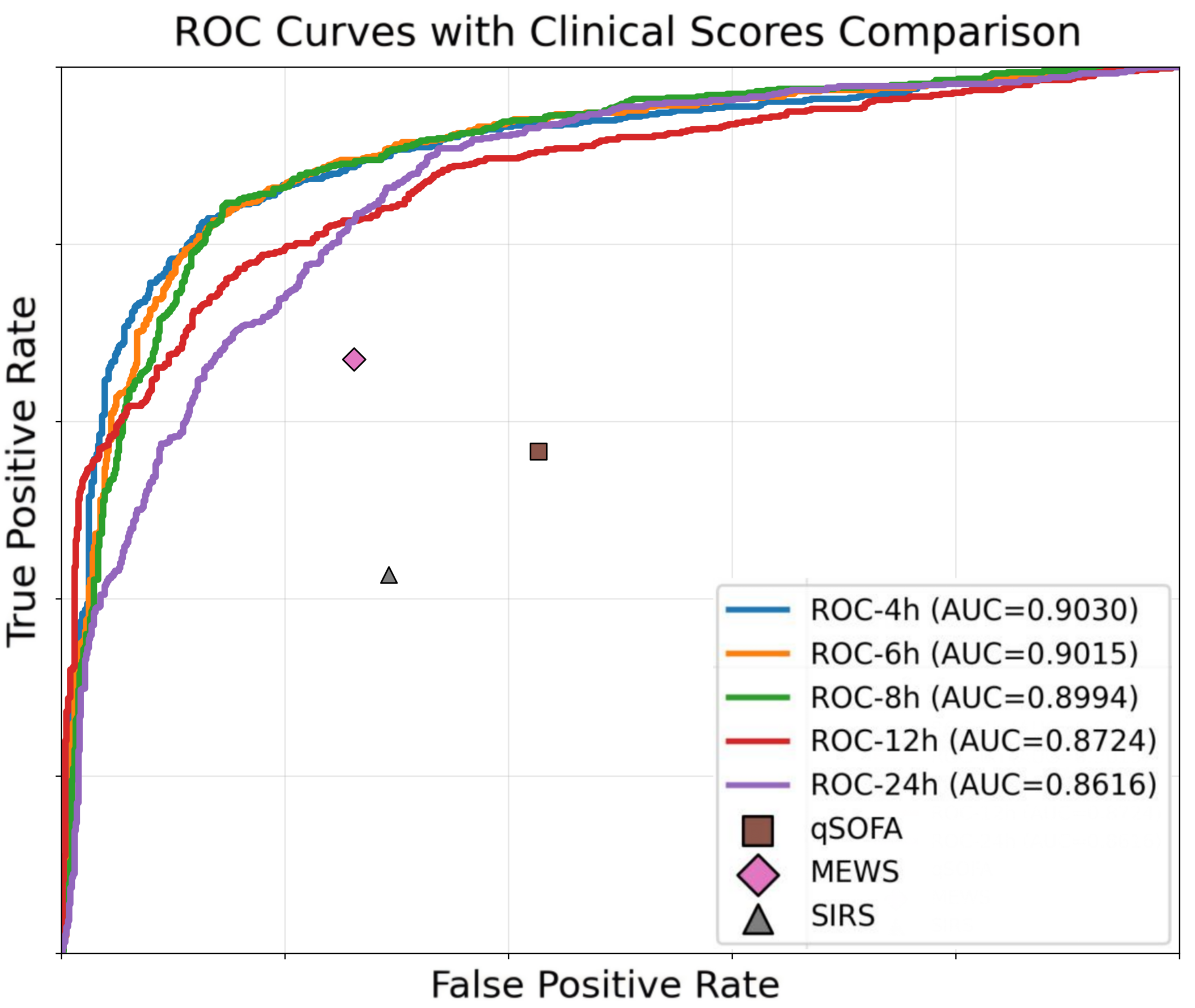}
    \caption{The receiver operating characteristic (ROC) curves for five early prediction tasks were analyzed at 24, 12, 8, 6, and 4 hours prior to disease onset. The predictive performance of the model for each task improves as the ROC curve approaches the upper left corner of the graph and demonstrates a larger area under the curve (AUC). The specific AUC values can be obtained from TABLE 1, while the graph also indicates the true positive rate and false positive rate corresponding to conventional physician diagnostic criteria for sepsis.}
    \label{fig:result}
\end{figure}
Our research involves five sequential prediction tasks for early sepsis detection at 24, 12, 8, 6, and 4 hours prior to onset. The specific results are presented in Figure. 5, including both the false discovery rate and true positive rate of the model's predictions alongside its ROC curve. Moreover, the outcomes include three widely used clinical sepsis assessment criteria (SIRS, qSOFA, and MEWS). As can be seen from the figure, compared with other criteria, MEWS shows relatively better performance. Furthermore, we also tested these three assessment criteria in five sequential prediction tasks. The outcomes are shown in Table 1. Because these criteria depend on short-term retrospective indicators and limited variable selection, they performed poorly in the aforementioned five prediction tasks. To further validate the superiority of our model, we conducted comparative experiments with nine baseline methods. These methods can be divided into three categories.

Firstly, we utilized three machine learning methods as follows: Support Vector Machine (SVM). This is a supervised learning algorithm based on maximum-margin classification principles, particularly effective for high-dimensional data classification; K-Nearest Neighbors (KNN). A lazy learning algorithm that classifies samples based on distance metrics to neighboring instances. This method is sensitive to local patterns. Based on the cross-validation, its optimal k-value is 7; Logistic Regression (LR). This method is a linear classifier, using the sigmoid function to model probabilities. Furthermore, it is widely adopted for binary classification tasks, such as our early sepsis prediction task.

Next, we selected six existing deep learning models for early sepsis prediction and reproduced their methods: Long Short-Term Memory (LSTM). This model has a recurrent neural network with gating mechanisms to address vanishing gradients and performs well in time series data prediction \cite{bib19}. Our experiments used a time series of length 31 with 47 variables, and the total number of parameters was 0.9M; Residual Network (ResNet). A CNN architecture, incorporating skip connections \cite{bib45}. This method can mitigate degradation in deep networks and enhance prediction accuracy. The configuration of the experiment is the same as above. The total number of parameters we used was 0.8M; Transformer. Based on the self-attention, this model is adept at capturing long-range dependencies. Furthermore, through its positional encoding and self-attention mechanism, this model can perform predictions on time-series data. In this experiment, we used 12 attention heads with 1.2M total parameters; Time-Phased Model. This method incorporates time-window partitioning to optimize periodic feature extraction for temporal analysis \cite{bib20}. By referring to previous related work, the total number of parameters we used was 1.1M; MGP-AttTCN \cite{bib21}. A temporal convolutional network, integrating multi-scale Gaussian processes and attention mechanisms. This method can process and predict complex time-series data. The total number of parameters we used was 1.2M; CNN-LSTM \cite{bib22}. This method combines convolutional and LSTM layers to jointly extract spatial and temporal features. The total number of parameters we used was 1.2M.

Finally, we utilized three clinical sepsis criteria \cite{bib25}, including \textbf{SIRS}, \textbf{qSOFA}, and \textbf{MEWS}. 

\begin{table}[t!]
    \centering
    \scriptsize
    \setlength{\abovecaptionskip}{0pt}
    \setlength{\tabcolsep}{5pt} 
    \caption{Performance Comparison of Different Models}
    \renewcommand{\arraystretch}{1.2}
    \begin{tabular}{l ccccccccccc} 
        \hline
        \multirow{2}{*}{Models} & 
        \multicolumn{2}{c}{24} & \multicolumn{2}{c}{12} & \multicolumn{2}{c}{8} & \multicolumn{2}{c}{6} & \multicolumn{2}{c}{4} \\
        & AUC & F1 & AUC & F1 & AUC & F1 & AUC & F1 & AUC & F1 \\ 
        \hline
        SVM & 0.582 & 0.573 & 0.741 & 0.711 & 0.766 & 0.734 & 0.803 & 0.744 & 0.825 & 0.766 \\
        KNN & 0.596 & 0.584 & 0.749 & 0.708 & 0.779 & 0.742 & 0.799 & 0.735 & 0.835 & 0.785 \\
        LR & 0.642 & 0.583 & 0.825 & 0.786 & 0.832 & 0.791 & 0.854 & 0.816 & 0.874 & 0.833 \\
        \hline
        LSTM & 0.782 & 0.744 & 0.856 & 0.819 & 0.873 & 0.833 & \textbf{0.884} & 0.846 & 0.891 & 0.852 \\
        Resnet & 0.653 & 0.629 & 0.763 & 0.724 & 0.776 & 0.742 & 0.781 & 0.744 & 0.829 & 0.811 \\
        Transformer & 0.822 & 0.774 & \textbf{0.866} & \textbf{0.833} & 0.871 & 0.845 & 0.883 & 0.849 & 0.894 & 0.863 \\
        Time-phased & 0.814 & 0.776 & 0.829 & 0.799 & 0.842 & 0.818 & 0.854 & 0.821 & 0.877 & 0.830 \\
        MGP-AttTCN & 0.826 & 0.765 & 0.865 & 0.821 & \textbf{0.876} & \textbf{0.849} & 0.883 & \textbf{0.859} & 0.894 & 0.863 \\
        CNN-LSTM & \textbf{0.845} & \textbf{0.792} & 0.861 & 0.815 & 0.874 & 0.843 & 0.882 & 0.853 & \textbf{0.900} & 0.864 \\
        \hline
        SIRS & 0.574 & 0.467 & 0.579 & 0.463 & 0.576 & 0.475 & 0.582 & 0.462 & 0.600 & 0.479 \\
        qSOFA & 0.564 & 0.435 & 0.551 & 0.416 & 0.546 & 0.445 & 0.551 & 0.437 & 0.575 & 0.432 \\
        MEWS & 0.672 & 0.492 & 0.676 & 0.498 & 0.719 & 0.525 & 0.712 & 0.506 & 0.728 & 0.532 \\
        \hline
        Ours & \textbf{0.861} & \textbf{0.827} & \textbf{0.872} & \textbf{0.835} & \textbf{0.899} & \textbf{0.862} & \textbf{0.901} & \textbf{0.865} & \textbf{0.903} & \textbf{0.871} \\
        \hline
    \end{tabular}
    \label{tab:performance_comparison}
\end{table}

All of the comparative results are summarized in Table 1. Notably, compared with the baseline methods, our model achieves a comprehensive improvement in all tasks. Moreover, as the prediction window approaches sepsis onset, our model’s performance also has continuous improvement (AUC improves from 0.861 at 24 hours to 0.903 at 4 hours). In all of the baseline methods, traditional machine learning models perform poorly. This result indicates that simple classifiers are limited in capturing complex spatiotemporal relationships for high-dimensional early sepsis prediction tasks. While LSTM and MGP-AttTCN obtain moderate results, they still have limitations in comprehensively extracting spatiotemporal features in patients’ data. Additionally, the CNN-LSTM model has shown the closest performance to ours at the 4-hour task. This method can deeply mine the complex relationships in these time series. However, in the real world, clinical text often contains some vital information \cite{bib26}. We cannot ignore the impact of these patients' text data. Therefore, this method lacks the utilization of textual patient data. Moreover, combined with the outcomes in Figure. 5, MEWS shows a better performance in TPR, while the qSOFA shows better performance in FPR. These results demonstrate that MEWS has a good performance in the sensitivity of sepsis testing. Moreover, qSOFA is more conservative in the testing of sepsis. In summary, our method has shown better performance compared with existing methods. We will demonstrate the success of our prior prediction module in the next part.

\subsection {Prior Predictive Performance}

\begin{figure}[htp]
    \centering
    \includegraphics[width=9cm]{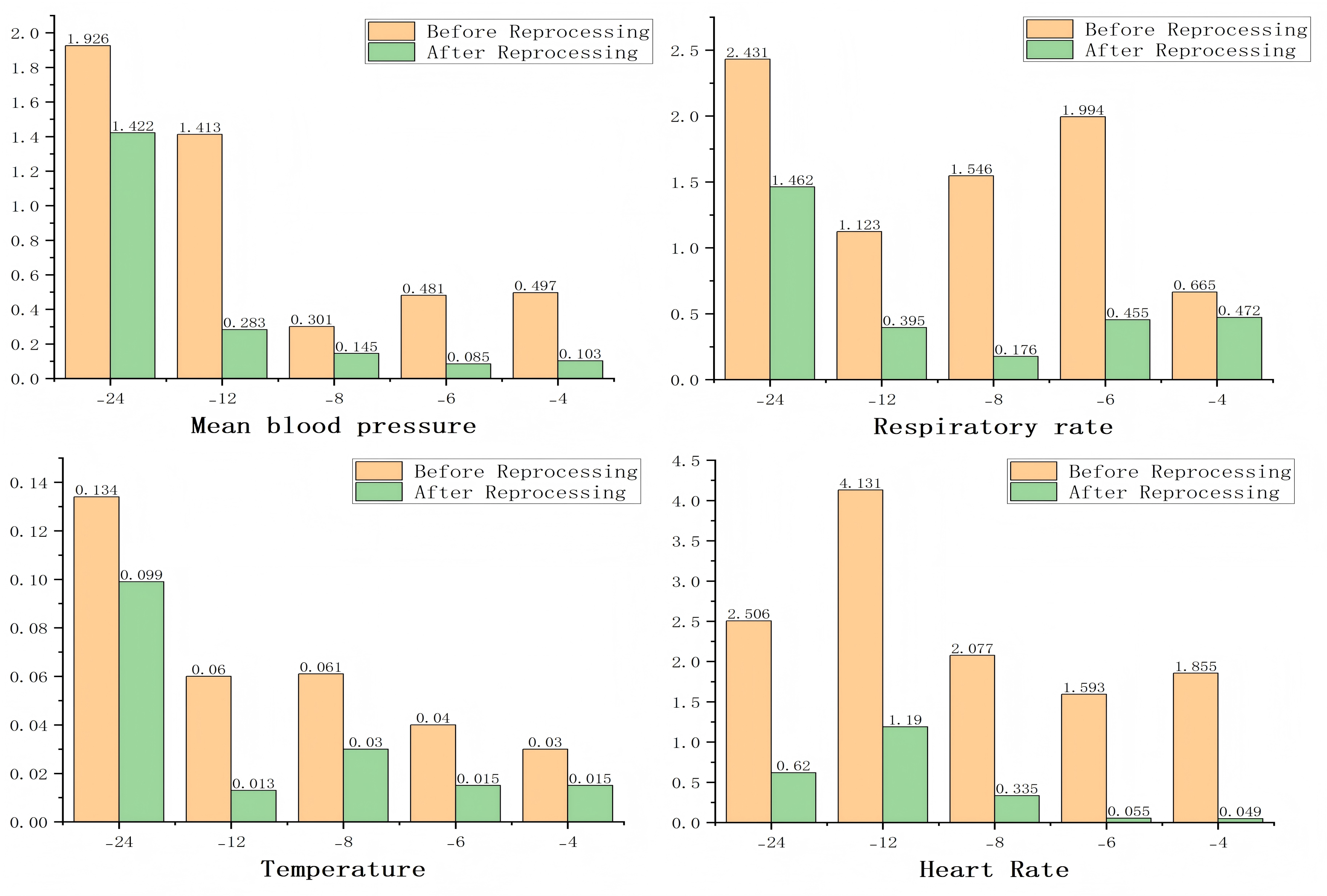}
    \caption{The figure shows the prediction results of some variables. The yellow part represents the prior prediction MSE without using the AI-driven post-processing component, while the green part represents the MSE after using this component. The time points of the task remain -24, -12, -8, -6, and -4.}
    \label{fig:result}
\end{figure}

\begin{table*}[htbp]
    \centering
    \scriptsize
    \setlength{\abovecaptionskip}{0pt}
    \caption{MSE of Partial Various Physiological Indicators After Reprocessing}
    \renewcommand{\arraystretch}{1.5}

    \setlength{\tabcolsep}{2pt} 
    
    \begin{tabular}{
        @{}
        >{\centering\arraybackslash}p{0.9cm}
        >{\centering\arraybackslash}p{0.9cm}
        >{\centering\arraybackslash}p{0.9cm}
        >{\centering\arraybackslash}p{0.9cm}
        >{\centering\arraybackslash}p{0.9cm}
        >{\centering\arraybackslash}p{0.9cm}
        >{\centering\arraybackslash}p{0.9cm}
        >{\centering\arraybackslash}p{0.9cm}
        >{\centering\arraybackslash}p{0.9cm}
        >{\centering\arraybackslash}p{0.9cm}
        >{\centering\arraybackslash}p{0.9cm}
        @{}
    }
        \hline
        Time & HR & SBP & DBP & MBP & RR & SPO\textsubscript{2} & Temp & Glu & GCS & ABPm \\
        \hline
        -24 & 0.620 & 1.373 & 0.858 & 1.422 & 1.462 & 1.543 & 0.099 & 0.372 & 0.862 & 1.023 \\
        -12 & 0.579 & 0.874 & 0.651 & 0.701 & 0.665 & 1.405 & 0.013 & 0.211 & 0.543 & 0.897 \\
        -8  & 1.190 & 0.583 & 0.503 & 0.283 & 1.239 & 0.542 & 0.030 & 0.210 & 0.449 & 0.841 \\
        -6  & 0.497 & 0.294 & 0.499 & 0.432 & 0.455 & 0.399 & 0.015 & 0.172 & 0.453 & 0.701 \\
        -4  & 0.506 & 0.322 & 0.483 & 0.316 & 0.472 & 0.403 & 0.015 & 0.128 & 0.430 & 0.625 \\
        \hline
    \end{tabular}
    \label{tab:mse_indicators}
\end{table*}

To demonstrate the accuracy and effectiveness of our prior prediction module, we select MSE as the metric. We present the partial prediction results in Table 2. Analyzing these outcomes, we can observe that most variables’ errors are kept below 1. This demonstrates that our prior prediction module has excellent predictive performance. However, for variables such as HR, SBP, MBP, RR, and SPO\textsubscript{2}, there are some time points where the prediction MSE exceeds 1. The reason for these larger errors is caused by the significant and complex variability inherent in these variables themselves.

\begin{figure}[htp]
    \centering
    \includegraphics[width=9cm]{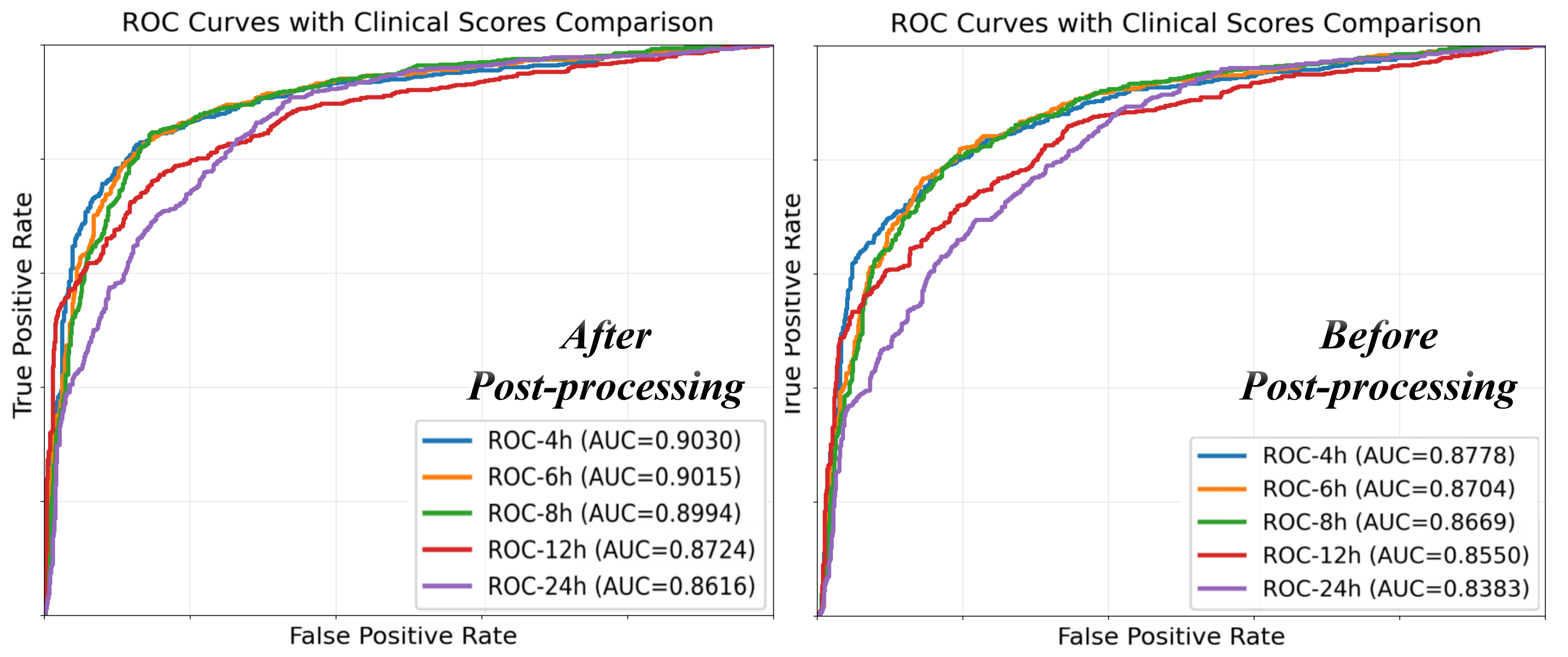} 
    \caption{The receiver operating characteristic (ROC) curves corresponding to the models with and without the post-processing module are displayed from left to right.}
    \label{fig:c}  
\end{figure}
To further demonstrate the accuracy and effectiveness of the post-processing component in our prior prediction module, we compared the results with and without this component. Figure.6 presents the results of partial data after prediction. The yellow part represents the variable errors without using the AI-driven post-processing component, while the green part does with it. From the figure, we can observe that the prediction errors of the data significantly decrease after using this component. Because the post-processing component can impose secondary constraints on some larger prediction errors, the result can avoid extreme prediction cases and enhance the accuracy of data prediction. Figure.7 demonstrates the impact of the component on subsequent classification predictions. It can be observed from the figure that after using the post-processing component, the average AUC increases by approximately 0.03. This may be due to the improved quality of the predicted data after processing, and it also avoids the impact of some extreme predicted values on the model's classification prediction during training. 

In summary, we have successfully demonstrated the excellent performance of this component.
The outputs of this module are not only part of the model’s prediction process but also serve as explanatory evidence that connects quantitative predictions with underlying physiological mechanisms.
By revealing which vital signs contribute most to the early warning signal and how their simulated trajectories evolve before sepsis onset, the module enables clinicians to interpret the model’s decision in physiologically meaningful terms.
Such interpretability facilitates trust, supports diagnostic reasoning, and helps physicians identify actionable early interventions rather than relying on opaque predictions.

In subsequent experiments, we will further demonstrate the impact of the prior prediction module on the overall performance of the model through ablation experiments.

\subsection {Performance Comparison of Various LLMs}

To further analyze the impact of different LLMs’ capabilities on our model’s predictive performance, we designed a comparative experiment based on six distinct local LLMs. The models we used are as follows: Bert, a general-domain pretrained model utilizing bidirectional Transformer architecture, demonstrates competent performance in short-text comprehension tasks \cite{bib40}. However, it has a limitation in maximum token length (512); Bio\_ClinicalBert. This is a Bert variant optimized for biomedical clinical texts \cite{bib42}. Compared with Bert, it shows better performance in medical terminology understanding; BioGPT. Based on GPT architecture, this model is a Microsoft-developed biomedical generative model \cite{bib23}. Compared with Bert, BioGPT supports longer clinical text inputs up to 1024 tokens; GPT2, a general generative pretrained model exhibiting superior contextual modeling capabilities and demonstrating intermediate-scale performance advantages over Bert variants in this study \cite{bib43}; Deepseek-R1 \cite{bib52}. This model is a large-scale model specializing in long-sequence modeling with long context support. In this experiment, we select 7B parameters; Mistral-7B, a 7B-parameter open-source efficient model, adopts a sliding window attention mechanism \cite{bib46}. It has proven to achieve optimal predictive performance while maintaining low computational costs.

\begin{table*}[htbp]
    \centering
    \scriptsize
    \setlength{\abovecaptionskip}{0pt}
    \caption{Performance Metrics (AUC and F1) of Different LLMs Under Various Parameter Settings}
    \renewcommand{\arraystretch}{1.5}

    \setlength{\tabcolsep}{3pt} 

    \begin{tabular}{
        @{} 
        >{\centering\arraybackslash}p{2cm}
        *{10}{>{\centering\arraybackslash}p{0.9cm}}
        @{} 
    }
        \hline
        \multirow{2}{*}{LLM} &
        \multicolumn{2}{c}{24} &
        \multicolumn{2}{c}{12} &
        \multicolumn{2}{c}{8} &
        \multicolumn{2}{c}{6} &
        \multicolumn{2}{c}{4} \\
        & AUC & F1 & AUC & F1 & AUC & F1 & AUC & F1 & AUC & F1 \\
        \hline
        Bert        & 0.732 & 0.698 & 0.755 & 0.721 & 0.774 & 0.735 & 0.788 & 0.750 & 0.797 & 0.764 \\
        Bio\_Bert   & 0.743 & 0.701 & 0.769 & 0.732 & 0.781 & 0.742 & 0.792 & 0.756 & 0.805 & 0.773 \\
        BioGPT      & 0.805 & 0.730 & 0.815 & 0.762 & 0.833 & 0.789 & 0.840 & 0.815 & 0.843 & 0.819 \\
        GPT2        & 0.812 & 0.758 & 0.830 & 0.775 & 0.846 & 0.806 & 0.852 & 0.829 & 0.863 & 0.842 \\
        Deepseek-R1 & 0.855 & 0.823 & 0.864 & 0.821 & 0.887 & 0.853 & 0.892 & 0.861 & 0.897 & 0.866 \\
        Mistral-7B  & 0.861 & 0.827 & 0.872 & 0.835 & 0.899 & 0.862 & 0.901 & 0.865 & 0.903 & 0.871 \\
        \hline
    \end{tabular}
    \label{tab:llm_performance}
\end{table*}

The study employs the same prediction task as mentioned above and shows the final outcomes in Table 3. Because our method integrates the text and extracted features together and inputs them into the LLM, it is a burden on the capacity of the prompt text of the model. The maximum token of Bert and Bio\_ClinicalBert is 512, which cannot support the complete input of the text. Therefore, compared with the other LLMs, Bert and Bio\_ClinicalBert have performed poorly in the sepsis prediction task. For BioGPT and GPT2, their maximum token have increased to 1024. It means that these models can contain most of the patient text, after allocating half of the token for processing the extracted features. Notably, the locally implemented Deepseek-R1 and Mistral-7B models show the best performance in our tasks. These models have the largest parameter sizes among selected models and maximum token lengths. Their tokens are 4096, which can well accommodate all the patients’ texts. Furthermore, during the training process, their parameter sizes can better extract spatiotemporal features. However, larger parameters mean occupying more device resources. For balancing training resources and model efficacy, we adopted a strategy of freezing most LLM layers while fine-tuning only selected layers. In summary, based on the outcomes of our experiment, we have shown the impact of choosing different LLMs on the results. In the subsequent ablation experiments, we will demonstrate the influence of various parameters on the prediction.

\subsection{Performance Comparison of eICU Database}
\begin{table*}[t!]
    \centering
    \scriptsize
    \setlength{\abovecaptionskip}{0pt}
    \setlength{\tabcolsep}{5pt}
    \caption{Performance Comparison of Different Models Based on eICU Dataset}
    \renewcommand{\arraystretch}{1.2}
    \begin{tabular}{l ccccccccccc} 
        \hline
        \multirow{2}{*}{Models} & 
        \multicolumn{2}{c}{24} & \multicolumn{2}{c}{12} & \multicolumn{2}{c}{8} & \multicolumn{2}{c}{6} & \multicolumn{2}{c}{4} \\
        & AUC & F1 & AUC & F1 & AUC & F1 & AUC & F1 & AUC & F1 \\ 
        \hline
        SVM & 0.587 & 0.553 & 0.734 & 0.706 & 0.742 & 0.714 & 0.783 & 0.745 & 0.811 & 0.779 \\
        KNN & 0.593 & 0.561 & 0.730 & 0.701 & 0.744 & 0.709 & 0.786 & 0.749 & 0.826 & 0.792 \\
        LR & 0.662 & 0.629 & 0.785 & 0.749 & 0.816 & 0.783 & 0.839 & 0.806 & 0.854 & 0.821 \\
        \hline
        LSTM & 0.779 & 0.743 & 0.850 & 0.821 & 0.861 & 0.836 & 0.876 & 0.842 & 0.890 & 0.859 \\
        Resnet & 0.694 & 0.655 & 0.756 & 0.733 & 0.767 & 0.732 & 0.779 & 0.751 & 0.813 & 0.785 \\
        Transformer & 0.821 & 0.790 & \textbf{0.851} & 0.820 & 0.859 & 0.825 & 0.875 & 0.849 & 0.889 & 0.854 \\
        Time-phased & 0.813 & 0.779 & 0.824 & 0.795 & 0.841 & 0.819 & 0.863 & 0.842 & 0.880 & 0.851 \\
        MGP-AttTCN & 0.825 & 0.796 & 0.850 & \textbf{0.823} & \textbf{0.863} & \textbf{0.840} & \textbf{0.880} & \textbf{0.851} & 0.893 & 0.858 \\
        CNN-LSTM & \textbf{0.832} & \textbf{0.806} & 0.844 & 0.816 & 0.861 & 0.836 & 0.878 & 0.846 & \textbf{0.895} & \textbf{0.860} \\
        \hline
        SIRS & 0.581 & - & 0.577 & - & 0.572 & - & 0.591 & - & 0.598 & - \\
        qSOFA & 0.596 & - & 0.581 & - & 0.572 & - & 0.596 & - & 0.602 & - \\
        MEWS & 0.693 & - & 0.693 & - & 0.684 & - & 0.703 & - & 0.711 & - \\
        \hline
        Ours & \textbf{0.852} & \textbf{0.831} & \textbf{0.874} & \textbf{0.840} & \textbf{0.887} & \textbf{0.855} & \textbf{0.894} & \textbf{0.860} & \textbf{0.901} & \textbf{0.863} \\
        \hline
    \end{tabular}
    \label{tab:eicu_performance}
\end{table*}

To further verify the validity and generalizability of our model’s results, we select eligible patient data from another database\cite{bib49}. The database used in our research is the eICU Collaborative Research Database (eICU-CRD v2.0) \cite{bib35}. This publicly available medical database is derived from Philips Healthcare's eICU telehealth program. It contains high-granularity data for over 200,000 intensive care unit (ICU) admissions across 208 hospitals in the United States from 2014 to 2015. Each patient has a unique identifier as "uniquePid," each hospitalization record has a "patientHealthSystemStayId," and each ICU stay has a "patientUnitStayId." By utilizing these identifiers, researchers can build cross-table data associations and accurately extract target data. 

The patient selection criteria are consistent with those applied to the MIMIC database in Section 3. After excluding the records with high missing rates, we finally gained 1513 positive cases and 2975 negative cases from the eICU database. Subsequently, we test the performance of the model on a new dataset. The results are presented in Table 4. By observation, it can be seen that the model performance at the final time point is comparable to the results tested on the MIMIC dataset. However, as the prediction time point shifted earlier, the accuracy of the predictions decreased at a slightly faster rate than observed in the previous experiment. Furthermore, building on the foundation of the prior experiment, we also test the performance of other methods for comparison. All of the prediction results are organized into the same table. Compared with these models, it is evident that our proposed model demonstrates better predictive performance across all prediction time points.

\subsection {Ablation Experiment}
\begin{table*}[htbp]
    \scriptsize
    \centering
    \setlength{\abovecaptionskip}{0pt}
    \caption{AUC Values for Different Model Components and Settings}

    \setlength{\tabcolsep}{1.5pt} 
    \renewcommand{\arraystretch}{1.5}

    \begin{tabular}{
        @{} 
        >{\centering\arraybackslash}m{1cm}
        >{\centering\arraybackslash}m{2cm}
        >{\centering\arraybackslash}m{2cm}
        >{\centering\arraybackslash}m{2.2cm}
        >{\centering\arraybackslash}m{1cm}
        >{\centering\arraybackslash}m{1cm}
        >{\centering\arraybackslash}m{1cm}
        >{\centering\arraybackslash}m{1cm}
        >{\centering\arraybackslash}m{1cm}
        @{}
    }
        \hline
        \multicolumn{4}{c}{Components} & \multicolumn{5}{c}{AUC Values} \\
        Patient Summary & Patching & Prior Prediction & Spatiotemporal Feature Extraction & 24h & 12h & 8h & 6h & 4h \\
        \hline
         & $\checkmark$ & $\checkmark$ & $\checkmark$ & 0.502 & 0.543 & 0.510 & 0.519 & 0.534 \\
        $\checkmark$ &  & $\checkmark$ & $\checkmark$ & 0.832 & 0.838 & 0.874 & 0.880 & 0.886 \\
        $\checkmark$ & $\checkmark$ &  & $\checkmark$ & 0.823 & 0.828 & 0.858 & 0.859 & 0.863 \\
        $\checkmark$ & $\checkmark$ & $\checkmark$ &  & 0.800 & 0.830 & 0.831 & 0.831 & 0.834 \\
        $\checkmark$ & $\checkmark$ & $\checkmark$ & $\checkmark$ & 0.861 & 0.872 & 0.899 & 0.901 & 0.903 \\
        \hline
    \end{tabular}
    \label{tab:auc_results}
\end{table*}

To further evaluate the contribution of each module to total performance, we designed ablation experiments through component adjustment. The detailed experimental results are summarized in Table 5. Notably, when we removed the patient summary, the model’s AUC value significantly dropped to approximately 0.5. This phenomenon indicates that the model lost its ability to make judgments for sepsis classification. Moreover, it demonstrates the critical importance of constructing appropriate and effective prompts in using LLMs for direct prediction. After removing the patch module, the whole time-series data were fed entirely into the model for prediction, which increased the complexity of matching between raw text prototypes and temporal data chunks. It caused the average AUC to decrease by 0.025 and showed the importance of patching. Furthermore, when the prior prediction module was disabled, the whole workflow of the model was reduced to a standalone classification model. Without having data on the predicted time point, the average AUC decreased by approximately 0.04. It demonstrates that the prior prediction module not only enhances the classification performance but also provides interpretability for the final outcomes. Finally, we removed the spatiotemporal extraction module from the overall framework. Through this process, the model relied solely on prompt engineering for prediction. LLMs would summarize the data chunks into text and combine with the patient summary to predict outcomes. The results show an average AUC decline of about 0.072, which highlights both the exceptional semantic comprehension capabilities of LLMs and the pivotal role of the spatiotemporal extraction module in the overall framework. These results prove the indispensable importance of each module to the whole process of the model. Moreover, these findings emphasize the distinct advantages of LLMs in processing textual data for constructing sepsis prediction models.

\subsection {Discussing the Importance of Observed Variables}
To analyze the variables influencing sepsis, this study utilized the Grad-CAM method to generate a grayscale map. The horizontal axis represents the time series within the 24 hours preceding sepsis onset, while the vertical axis lists multiple clinical variables \cite{bib27}. The intensity of grayscale reflects the contribution of each variable to the sepsis prediction model at different time points. The darker the color, the greater the contribution of that variable to the model's prediction of sepsis at that time point; conversely, the lighter the color, the smaller the contribution of that variable.
\begin{figure}
    \centering
    \includegraphics[width=5cm]{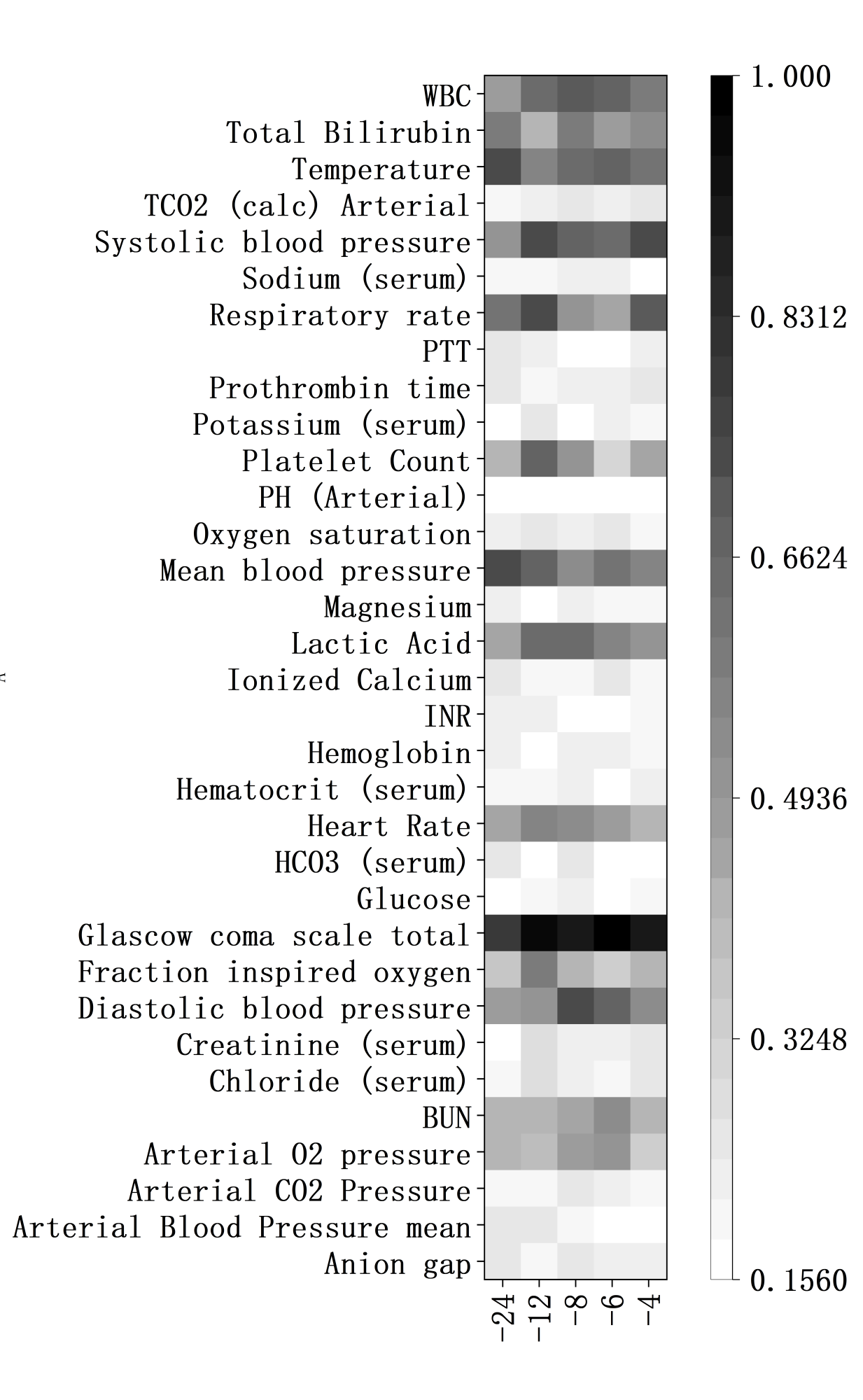}
    \caption{The degree of influence of different variables on the results. The vertical axis shows variable names, horizontal axis indicates prediction time points.}
    \label{fig:var_appendix}
\end{figure}
As shown in Figure.8, multiple clinical variables make significant contributions to sepsis prediction\cite{bib41}, particularly as sepsis approaches. For example, white blood cell count (WBC), total Bilirubin, temperature, systolic blood pressure, respiratory rate, platelet count, mean blood pressure, lactic acid, heart rate, Glasgow coma scale total, fraction of inspired oxygen, diastolic blood pressure, BUN, and arterial O2 pressure, among others, demonstrate strong contributions in the hours leading up to sepsis onset. These variables are not only closely related to the physiological mechanisms of sepsis \cite{bib28} but also highly consistent with multiple components of the SOFA score (such as respiratory, coagulation, liver function, cardiovascular function, etc.).

White blood cell count is typically a marker of infection \cite{bib29}, and an increase in white blood cells is a common clinical phenomenon in the early stages of sepsis. Figure.8 shows that WBC contributes significantly to the model's prediction in the hours leading up to sepsis onset, indicating its important role in identifying sepsis \cite{bib30}. Elevated bilirubin reflects liver dysfunction, which is closely associated with the common liver damage observed in sepsis patients. The contribution of total bilirubin in Figure.8 is relatively high, consistent with its clinical importance in the diagnosis of sepsis. Changes in body temperature are a common indicator of sepsis, especially high fever. In Figure.8, changes in body temperature show a strong contribution prior to the onset of sepsis, consistent with the clinical emphasis on monitoring body temperature. Hypotension is a typical manifestation of sepsis, reflecting cardiovascular system dysfunction. In Figure.8, systolic pressure and mean pressure \cite{bib31} are two variables with significant contributions to the model, indicating their important diagnostic value in the early stages of sepsis. Elevated lactate levels typically reflect tissue hypoxia and metabolic acidosis, which are common physiological changes in sepsis patients. The contribution of lactate changes to sepsis prediction in Figure.8 is also significant, highlighting its role as an important physiological marker. The Glasgow Coma Scale reflects central nervous system function \cite{bib32}, and sepsis may cause altered mental status or coma. Its high contribution in Figure.8 indicates that changes in mental status are an important indicator of sepsis.

However, certain key variables have relatively low contributions in Figure.8. For example, variables such as creatinine \cite{bib47}, although important in the diagnosis and assessment of sepsis, are not significantly reflected in the model. This may be due to the fact that in the early stages of sepsis, creatinine data are often missing due to low testing frequency, and the model cannot learn its early warning value from incomplete data. Additionally, early fluid resuscitation strategies may temporarily dilute creatinine, masking its rising trend, and the model cannot distinguish this treatment interference. This is also a limitation of this study, and future model improvements could place greater emphasis on these treatment variables to enhance the accuracy of early sepsis identification.

\begin{figure}[htp]
    \centering
    \includegraphics[width=9cm]{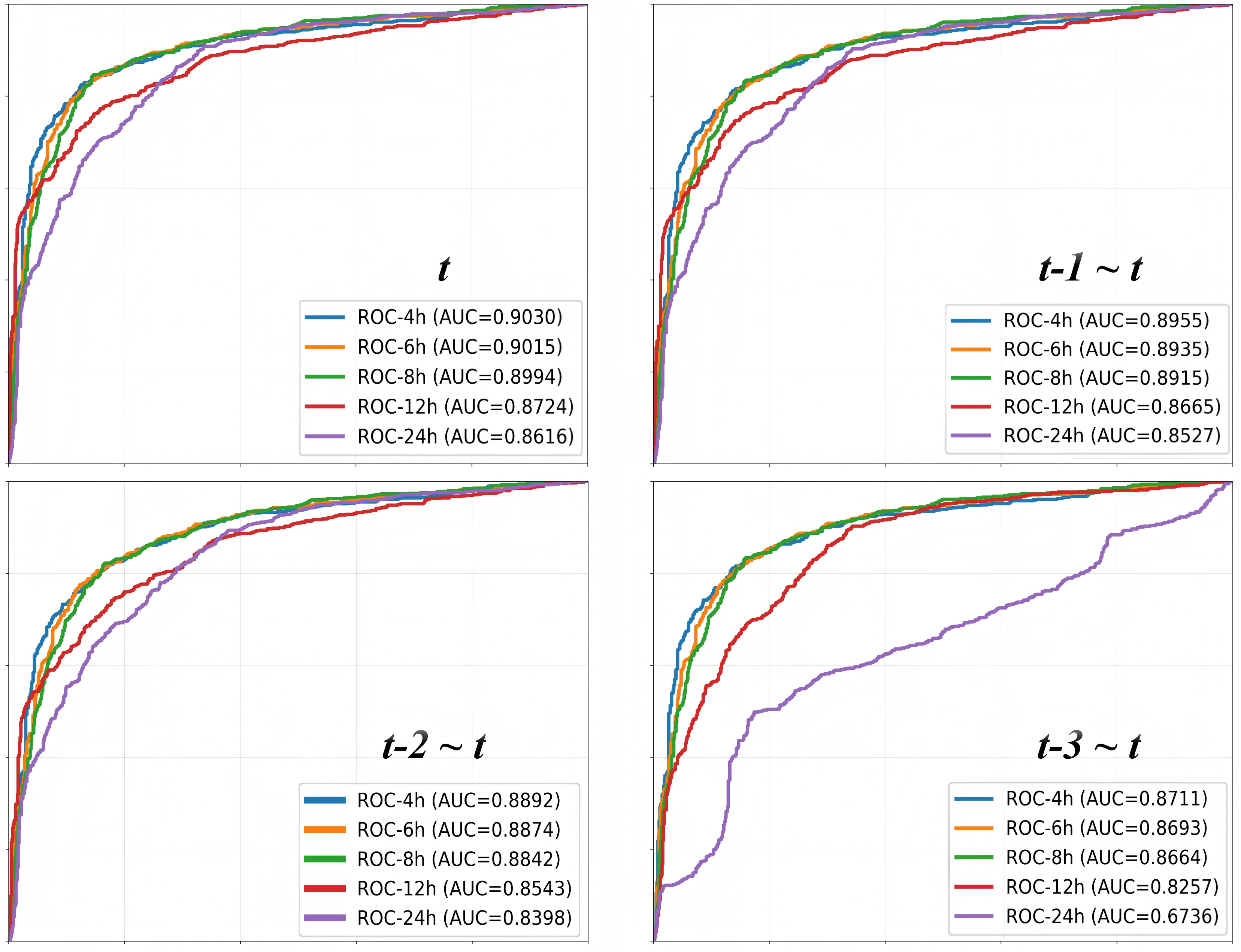}  
    \caption{From top-left to bottom-right, the figure sequentially displays the receiver operating characteristic (ROC) curves of the a priori prediction module for different prediction horizons: Task 1 (1-hour), Task 2 (2-hour), Task 3 (3-hour), and Task 4 (4-hour).}
    \label{fig:different} 
\end{figure}
\subsection {Parameter Experiment}
\textbf{Adjusting the prior prediction time length}. To illustrate the impact of this module on the final classification model’s predictions, we designed the experiment by adjusting the prediction time length. The final results are shown in Figure.9. Analyzing these outcomes, we observe that when the prediction length is set to less than 4 hours, the prediction results decrease gently, while the overall predictive performance of the model remains stable and effective. However, when the prediction length is extended to 4 hours, because the whole time span is 30 hours, the prediction performance for the 24-hour pre-onset period declines sharply. Additionally, while the other prediction tasks show performance decreases, the model still maintains strong predictive capability. This experiment has shown the impact of prediction time on the model’s effect and demonstrated the module's ability to maintain prediction stability across dynamic time windows.

\subsection{Discussion}
The proposed LLM-guided temporal simulation framework promises to boost reliability and interpretability in sepsis early warning systems. Unlike black-box AI, it simulates physiological indicator evolution before predictions, aligning with physicians’ reasoning—e.g., linking lactate, blood pressure, or oxygen trends to sepsis risk. This fosters clinician trust, cuts alarm fatigue, and improves ICU early intervention adherence.

Implementation is straightforward: integrates with hospital systems using routine EHR data (vitals, labs, notes), no extra costs or invasives. The prior-prediction module offers real-time trajectory visualizations for patient risk tracking, aiding triage and multidisciplinary decisions.

In the long term, it bridges data-driven AI with human expertise, shifting sepsis prediction to interactive, explainable support for smart ICUs. Future research will focus on multi-hospital validation, bedside UI tweaks, and trials on intervention outcomes and survival.

\section{CONCLUSION}
We proposed a novel LLM-Based Spatiotemporal Feature Extraction Model to predict physiological indicators and disease probability at the onset of sepsis. Compared with the baseline standards, our model showed a superior performance across a series of early prediction tasks. Furthermore, we designed subsequent experiments. The experimental results and analyses validated the predictive efficacy of our model. Notably, our model provides more interpretable physiological indicator predictions for final disease outcomes. These indicators can assist medical professionals in better analyzing the patients’ condition and then enable timely therapeutic interventions. We believe that the application of early prediction could significantly reduce sepsis mortality rates in the future. The current work suggests potential for further development in patient prompt design, and future improvements will focus on optimizing medical prompts as a primary objective.


\section{References}
\begin{thebibliography}{00}
\bibitem{bib1} M. Singer, C. S. Deutschman, C. W. Seymour, and et al., ``The third international consensus definitions for sepsis and septic shock (Sepsis-3),'' {\it JAMA}, vol. 315, no. 8, pp. 801--810, Jun. 2016, doi: 10.1001/jama.2016.0287.
\bibitem{bib2} M. Cecconi, L. Evans, M. Levy, and et al., ``Sepsis and septic shock,'' {\it Lancet}, vol. 392, no. 10141, pp. 75--87, Aug. 2018, doi: 10.1016/S0140-6736(18)30696-2.
\bibitem{bib3} T. van der Poll, F. L. Van De Veerdonk, B. P. Scicluna, and et al., ``The immunopathology of sepsis and potential therapeutic targets,'' {\it Nat. Rev. Immunol.}, vol. 17, no. 7, pp. 407--420, Jul. 2017, doi: 10.1038/nri.2017.36.
\bibitem{bib4} C. W. Seymour, V. X. Liu, T. J. Iwashyna, F. M. Brunkhorst, T. D. Rea, A. Scherag, and et al., ``Assessment of clinical criteria for sepsis: for the third international consensus definitions for sepsis and septic shock (Sepsis-3),'' {\it JAMA}, vol. 315, no. 8, pp. 762--774, Jun. 2016, doi: 10.1001/jama.2016.0288.
\bibitem{bib8} H. ElMoaqet, D. M. Tilbury, and S. K. Ramachandran, ``Multi-step ahead predictions for critical levels in physiological time series,'' {\it IEEE Trans. Cybern.}, vol. 46, no. 7, pp. 1704--1717, Jul. 2016, doi: 10.1109/TCYB.2016.2561974.
\bibitem{bib9} N. Oh, W. C. Cha, J. H. Seo, S. G. Choi, J. M. Kim, C. R. Chung, and et al., ``ChatGPT predicts in-hospital all-cause mortality for sepsis: In-context learning with the Korean Sepsis Alliance Database,'' {\it Healthcare Inform. Res.}, vol. 30, no. 3, pp. 266--276, Sep. 2024, doi: 10.4258/hir.2024.30.3.266.
\bibitem{bib10} P. E. Marik and A. M. Taeb, ``SIRS, qSOFA and new sepsis definition,'' {\it J. Thorac. Dis.}, vol. 9, no. 4, pp. 943, Apr. 2017, doi: 10.21037/jtd.2017.03.125.
\bibitem{bib12} A. E. Johnson, L. Bulgarelli, L. Shen, A. Gayles, A. Shammout, S. Horng, and et al., ``MIMIC-IV, a freely accessible electronic health record dataset,'' {\it Sci. Data}, vol. 10, no. 1, pp. 1, Jan. 2023, doi: 10.1038/s41597-022-01899-x.
\bibitem{bib13} J. E. Gotts and M. A. Matthay, ``Sepsis: pathophysiology and clinical management,'' {\it BMJ}, vol. 353, no. 1, 2016, doi: 10.1136/bmj.i1585.
\bibitem{bib63} J. K. Agor, R. Li, O. Y. ¨Ozaltın, ``Septic shock prediction and knowledge discovery through temporal pattern mining,'' {\it Artif. Intell. Med.}, vol. 132, pp. 102406, 2022, doi: 10.1016/j.artmed.2022.102406.
\bibitem{bib14} K.-H. Goh, L. Wang, A. Y. K. Yeow, H. Poh, K. Li, and et al., ``Artificial intelligence in sepsis early prediction and diagnosis using unstructured data in healthcare,'' {\it Nat. Commun.}, vol. 12, no. 1, pp. 711, Jan. 2021, doi: 10.1038/s41467-021-20910-4.
\bibitem{bib15} Q. Li, D. Li, W. Nie, H. Jiao, Z. Wu, and et al., ``Temporal and spatial analysis in early sepsis prediction via causal disentanglements,'' {\it IEEE Trans. Knowl. Data Eng.}, to be published, 2025.
\bibitem{bib53} M. Zhu, J. Xia, X. Jin, M. Yan, G. Cai, and et al., ``Class weights random forest algorithm for processing class imbalanced medical data,'' {\it IEEE Access}, vol. 6, pp. 4641--4652, Jan. 2018, doi: 10.1109/ACCESS.2018.2789429.
\bibitem{bib17} W. Sun, Z. Cai, Y. Li, F. Liu, S. Fang, and et al., ``Data processing and text mining technologies on electronic medical records: a review,'' {\it J. Healthcare Eng.}, vol. 2018, no. 1, pp. 4302425, Dec. 2018, doi: 10.1155/2018/4302425.
\bibitem{bib50} R. Dürichen, M. A. F. Pimentel, L. Clifton, et al., ``Multitask Gaussian processes for multivariate physiological time-series analysis,'' {\it IEEE Trans. Biomed. Eng.}, vol. 62, no. 1, pp. 314--322, Jan. 2015, doi: 10.1109/TBME.2014.2351376.
\bibitem{bib18} T. Desautels, J. Calvert, J. Hoffman, M. Jay, Y. Kerem, L. Shieh, and et al., ``Prediction of sepsis in the intensive care unit with minimal electronic health record data: a machine learning approach,'' {\it JMIR Med. Inform.}, vol. 4, no. 3, pp. e5909, Sep. 2016, doi: 10.2196/medinform.5909.
\bibitem{bib19} D. A. Kaji, J. R. Zech, J. S. Kim, S. K. Cho, N. S. Dangayach, A. B. Costa, and et al., ``An attention based deep learning model of clinical events in the intensive care unit,'' {\it PLOS ONE}, vol. 14, no. 2, pp. e0211057, Feb. 2019, doi: 10.1371/journal.pone.0211057.
\bibitem{bib20} X. Li, X. Xu, F. Xie, X. Xu, Y. Sun, X. Liu, and et al., ``A time-phased machine learning model for real-time prediction of sepsis in critical care,'' {\it Crit. Care Med.}, vol. 48, no. 10, pp. e884--e888, Oct. 2020, doi: 10.1097/CCM.0000000000004494.
\bibitem{bib46} M. Rosnati and V. Fortuin, ``MGP-AttTCN: an interpretable machine learning model for the prediction of sepsis,'' {\it PLOS ONE}, vol. 16, no. 5, pp. e0251248, May 2021, doi: 10.1371/journal.pone.0251248.
\bibitem{bib22} W. Zha, Y. Liu, Y. Wan, R. Luo, D. Li, S. Yang, and et al., ``Forecasting monthly gas field production based on the CNN-LSTM model,'' {\it Energy}, vol. 260, pp. 124889, Dec. 2022, doi: 10.1016/j.energy.2022.124889.
\bibitem{bib23} R. Luo, L. Sun, Y. Xia, T. Qin, S. Zhang, H. Poon, and et al., ``BioGPT: generative pre-trained transformer for biomedical text generation and mining,'' {\it Brief. Bioinform.}, vol. 23, no. 6, pp. bbac409, Dec. 2022, doi: 10.1093/bib/bbac409.
\bibitem{bib24} M. H. Tahan, M. Ghasemzadeh, and S. Asadi, ``A novel embedded discretization-based deep learning architecture for multivariate time series classification,'' {\it IEEE Trans. Ind. Inform.}, vol. 19, no. 4, pp. 5976--5985, Apr. 2023, doi: 10.1109/TII.2022.3188839.
\bibitem{bib25} A. Camacho-Gonzalez, P. W. Spearman, and B. J. Stoll, ``Neonatal infectious diseases: evaluation of neonatal sepsis,'' {\it Pediatr. Clin. North Am.}, vol. 60, no. 2, pp. 367, Apr. 2013, doi: 10.1016/j.pcl.2012.12.003.
\bibitem{bib51} A. Zhou, R. Beyah, and R. Kamaleswaran, ``OnAI-Comp: an online AI experts competing framework for early sepsis detection,'' {\it IEEE/ACM Trans. Comput. Biol. Bioinform.}, vol. 19, no. 6, pp. 3595--3603, Nov./Dec. 2022, doi: 10.1109/TCBB.2021.3122405.
\bibitem{bib26} M. Y. Yan, L. T. Gustad, and {\O}. Nytrø, ``Sepsis prediction, early detection, and identification using clinical text for machine learning: a systematic review,'' {\it J. Am. Med. Inform. Assoc.}, vol. 29, no. 3, pp. 559--575, Mar. 2022, doi: 10.1093/jamia/ocab270.
\bibitem{bib27} W. A. Knaus, X. Sun, P. O. Nystrom, and D. P. Wagner, ``Evaluation of definitions for sepsis,'' {\it Chest}, vol. 101, no. 6, pp. 1656--1662, Jun. 1992, doi: 10.1378/chest.101.6.1656.
\bibitem{bib28} A. M. Esper, M. Moss, C. A. Lewis, R. Nisbet, D. M. Mannino, and G. S. Martin, ``The role of infection and comorbidity: factors that influence disparities in sepsis,'' {\it Crit. Care Med.}, vol. 34, no. 10, pp. 2576--2582, Oct. 2006, doi: 10.1097/01.CCM.0000240646.29109.6A.
\bibitem{bib61} Y. Zhu, A. Mueen, and E. Keogh, Admissible time series motif discovery with missing data,'' {\it IEEE Trans. Knowl. Data Eng.}, vol. 33, no. 11, pp. 3402--3415, Nov. 2021, doi: 10.1109/TKDE.2019.2948196.
\bibitem{bib48} G. P. Castelli, C. Pognani, M. Cita, A. Stuani, L. Sgarbi, and R. Paladini, ``Procalcitonin, C-reactive protein, white blood cells and SOFA score in ICU: diagnosis and monitoring of sepsis,'' {\it Minerva Anestesiol.}, vol. 72, no. 1/2, pp. 69, Jan./Feb. 2006, doi: 10.1007/s12340-006-0012-6.
\bibitem{bib49} Z. Peng, J. S. Schouten, D. Silvertand, et al., ``External validation complexities: a comparative study of late-onset sepsis prediction models across multiple clinical environments,'' {\it IEEE Trans. Biomed. Eng.}, to be published, 2025, doi: 10.1109/TBME.2025.3618080.
\bibitem{bib30} B. S. Karon, N. V. Tolan, A. M. Wockenfus, D. R. Block, N. A. Baumann, S. C. Bryant, and C. M. Clements, ``Evaluation of lactate, white blood cell count, neutrophil count, procalcitonin and immature granulocyte count as biomarkers for sepsis in emergency department patients,'' {\it Clin. Biochem.}, vol. 50, no. 16-17, pp. 956--958, Oct. 2017, doi: 10.1016/j.clinbiochem.2017.06.010.
\bibitem{bib31} M. W. Dünser, J. Takala, H. Ulmer, V. D. Mayr, G. Luckner, S. Jochberger, and et al., ``Arterial blood pressure during early sepsis and outcome,'' {\it Intensive Care Med.}, vol. 35, pp. 1225--1233, Jul. 2009, doi: 10.1007/s00134-009-1544-3.
\bibitem{bib32} F. C. Reith, R. Van den Brande, A. Synnot, R. Gruen, and A. I. Maas, ``The reliability of the Glasgow Coma Scale: a systematic review,'' {\it Intensive Care Med.}, vol. 42, pp. 3--15, Jan. 2016, doi: 10.1007/s00134-015-4124-3.
\bibitem{bib47} K. Doi, P. S. Yuen, C. Eisner, X. Hu, A. Leelahavanichkul, and R. A. Star, ``Reduced production of creatinine limits its use as marker of kidney injury in sepsis,'' {\it J. Am. Soc. Nephrol.}, vol. 20, no. 6, pp. 1217--1221, Jun. 2009, doi: 10.1681/ASN.2008090955.
\bibitem{bib34} B. Meskó, ``Prompt engineering as an important emerging skill for medical professionals: tutorial,'' {\it J. Med. Internet Res.}, vol. 25, pp. e50638, Mar. 2023, doi: 10.2196/50638.
\bibitem{bib35} T. J. Pollard, A. E. W. Johnson, J. D. Raffa, L. A. Celi, R. G. Mark, and O. Badawi, ``The eICU Collaborative Research Database, a freely available multi-center database for critical care research,'' {\it Sci. Data}, vol. 5, no. 1, pp. 1--13, Jan. 2018, doi: 10.1038/sdata.2018.178.
\bibitem{bib36} I. J. Goodfellow, J. Shlens, C. Szegedy, and et al., ``Explaining and harnessing adversarial examples,'' {\it arXiv Prepr.}, 2014, doi: 10.48550/arXiv.1412.6572, [Online]. Available: https://arxiv.org/abs/1412.6572.
\bibitem{bib37} L. M. Fleuren, T. L. Klausch, C. L. Zwager, L. J. Schoonmade, T. Guo, L. F. Roggeveen, and et al., ``Machine learning for the prediction of sepsis: a systematic review and meta-analysis of diagnostic test accuracy,'' {\it Intensive Care Med.}, vol. 46, pp. 383--400, Feb. 2020, doi: 10.1007/s00134-019-05872-y.
\bibitem{bib65} H. Meng, L. Guo, Y. Pan, B. Kong, W. Shuai, H. Huang, ``Machine learning based clinical prediction model for 1-year mortality in Sepsis patients with atrial fibrillation,'' {\it Heliyon}, vol. 10, pp. e38730, 2024, doi: 10.1016/j.heliyon.2024.e38730.
\bibitem{bib59} D. Dera, S. Ahmed, N. C. Bouaynaya, and G. Rasool, TRustworthy uncertainty propagation for sequential time-series analysis in RNNs,'' {\it IEEE Trans. Knowl. Data Eng.}, vol. 36, no. 2, pp. 882--896, Feb. 2024, doi: 10.1109/TKDE.2023.3288628.
\bibitem{bib38} J.-L. Vincent, S. M. Opal, J. C. Marshall, and K. J. Tracey, ``Sepsis definitions: time for change,'' {\it Lancet}, vol. 381, no. 9868, pp. 774--775, Feb. 2013, doi: 10.1016/S0140-6736(12)61815-7.
\bibitem{bib39} M. Jin, S. Wang, L. Ma, Z. Chu, J. Y. Zhang, X. Shi, and et al., ``Time-LLM: time series forecasting by reprogramming large language models,'' {\it arXiv Prepr.}, 2023, doi: 10.48550/arXiv.2310.02307, [Online]. Available: https://arxiv.org/abs/2310.02307.
\bibitem{bib64} C. Düsing, P. Cimiano, S. Rehberg, C. Scherer, O. Kaup, C. Köster, S. Hellmich, D. Herrmann, K. L. Meier, S. Claßen, R. Borgstedt, ``Integrating federated learning for improved counterfactual explanations in clinical decision support systems for sepsis therapy,'' {\it Artif. Intell. Med.}, vol. 157, pp. 102982, 2024, doi: 10.1016/j.artmed.2024.102982.
\bibitem{bib40} Y. Liu, M. Ott, N. Goyal, J. Du, M. Joshi, D. Chen, and et al., ``RoBERTa: a robustly optimized BERT pretraining approach,'' {\it arXiv Prepr.}, 2019, doi: 10.48550/arXiv.1907.11692, [Online]. Available: https://arxiv.org/abs/1907.11692.
\bibitem{bib52} D. Guo, D. Yang, H. Zhang, J. Song, R. Zhang, R. Xu, and et al., ``Deepseek-r1: incentivizing reasoning capability in LLMs via reinforcement learning,'' {\it arXiv Prepr.}, 2025, doi: 10.48550/arXiv.2501.12948, [Online]. Available: https://arxiv.org/abs/2501.12948.
\bibitem{bib54} C. Sun, H. Li, M. Song, and S. Hong, ``A ranking-based cross-entropy loss for early classification of time series,'' {\it IEEE Trans. Neural Netw. Learn. Syst.}, vol. 35, no. 8, pp. 11194--11204, Aug. 2024, doi: 10.1109/TNNLS.2023.3250203.
\bibitem{bib42} E. Alsentzer, J. R. Murphy, W. Boag, W. H. Weng, D. Jin, T. Naumann, and et al., ``Publicly available clinical BERT embeddings,'' {\it arXiv Prepr.}, 2019, doi: 10.48550/arXiv.1904.03323, [Online]. Available: https://arxiv.org/abs/1904.03323.
\bibitem{bib43} V. Kieuvongngam, B. Tan, and Y. Niu, ``Automatic text summarization of COVID-19 medical research articles using BERT and GPT-2,'' {\it arXiv Prepr.}, 2020, doi: 10.48550/arXiv.2006.01997, [Online]. Available: https://arxiv.org/abs/2006.01997.
\bibitem{bib58} Q. Li, D. Li, W. Nie, H. Jiao, Z. Wu, and et al., Temporal and spatial analysis in early sepsis prediction via causal disentanglements,'' {\it IEEE Trans. Knowl. Data Eng.}, to be published, 2025, doi: 10.1109/TKDE.2024.3401849. 
\bibitem{bib44} B. Zhang, Z. Liu, C. Cherry, and O. Firat, ``When scaling meets LLM finetuning: the effect of data, model and finetuning method,'' {\it arXiv Prepr.}, 2024, doi: 10.48550/arXiv.2402.17193, [Online]. Available: https://arxiv.org/abs/2402.17193.
\bibitem{bib45} S. Targ, D. Almeida, and K. Lyman, ``Resnet in resnet: generalizing residual architectures,'' {\it arXiv Prepr.}, 2016, doi: 10.48550/arXiv.1603.08029, [Online]. Available: https://arxiv.org/abs/1603.08029.
\bibitem{bib21} G. Jung, M. A. Hiltunen, K. R. Joshi, R. D. Schlichting, and C. Pu, ``Mistral: Dynamically managing power, performance, and adaptation cost in cloud infrastructures,'' in {\it Proc. 2010 IEEE 30th Int. Conf. Distrib. Comput. Syst.}, Minneapolis, MN, USA, Jun. 2010, pp. 62--73, doi: 10.1109/ICDCS.2010.34.
\bibitem{bib33} L. Faes, K. H. Chon, and G. Nollo, ``A method for the time-varying nonlinear prediction of complex nonstationary biomedical signals,'' {\it IEEE Trans. Biomed. Eng.}, vol. 56, no. 2, pp. 206--209, Feb. 2009, doi: 10.1109/TBME.2008.2008726.
\bibitem{bib29} D. A. Birrenkott, M. A. F. Pimentel, P. J. Watkinson, and D. A. Clifton, ``A robust fusion model for estimating respiratory rate from photoplethysmography and electrocardiography,'' {\it IEEE Trans. Biomed. Eng.}, vol. 65, no. 9, pp. 2033--2041, Sep. 2018, doi: 10.1109/TBME.2017.2778265.
\bibitem{bib41} I. A. Rezek and S. J. Roberts, ``Stochastic complexity measures for physiological signal analysis,'' {\it IEEE Trans. Biomed. Eng.}, vol. 45, no. 9, pp. 1186--1191, Sep. 1998, doi: 10.1109/10.718287.
\bibitem{bib60} Y. Li, J. Li, Y. Li, and Q. Li, Time series anomaly detection with adversarial reconstruction networks,'' {\it IEEE Trans. Knowl. Data Eng.}, vol. 35, no. 10, pp. 10245--10258, Oct. 2023, doi: 10.1109/TKDE.2021.3137861. 
\bibitem{bib62} S. M. Lauritsen, M. E. Kaløra, E. L. Kongsgaard, K. M. Lauritsen, M. J. Jørgensen, J. Lange, B. Thiesson, ``Early detection of sepsis utilizing deep learning on electronic health record event sequences,'' {\it Artif. Intell. Med.}, vol. 104, pp. 101820, 2020, doi: 10.1016/j.artmed.2020.101820.


\end{thebibliography}
\end{document}